\documentclass[lettersize,journal]{IEEEtran}
\usepackage{amsmath,amsfonts}
\usepackage{algorithmic}
\usepackage{array}
\usepackage[caption=false,font=normalsize,labelfont=sf,textfont=sf]{subfig}
\usepackage{textcomp}
\usepackage{stfloats}
\usepackage{url}
\usepackage{verbatim}
\usepackage{graphicx}
\usepackage{cite}

\usepackage{caption}
\usepackage{url}
\usepackage[]{hyperref}         
\hypersetup{
hidelinks,
urlcolor = black
}
\usepackage{booktabs}
\usepackage{multirow}
\usepackage{xcolor}
\usepackage{xspace}
\usepackage{colortbl}
\usepackage[linesnumbered,ruled,vlined]{algorithm2e}
\usepackage{svg}
\usepackage{amssymb}

\def\BibTeX{{\rm B\kern-.05em{\sc i\kern-.025em b}\kern-.08em
    T\kern-.1667em\lower.7ex\hbox{E}\kern-.125emX}}
\usepackage{balance}
\usepackage{tikz}

\makeatletter
\DeclareRobustCommand\onedot{\futurelet\@let@token\@onedot}
\def\@onedot{\ifx\@let@token.\else.\null\fi\xspace}

\def\eg{\emph{e.g}\onedot} 

\def\ie{\emph{i.e}\onedot}

\makeatother

\newcommand{\tool}{\textsc{ArcGen}\xspace}

\usepackage{xargs} 
\newcommand{\change}[1]{{#1}}

\begin{document}

\title{\tool: Generalizing Neural Backdoor Detection Across Diverse Architectures}

\author{Zhonghao Yang, Cheng Luo, Daojing He, Yiming Li, and Yu Li

\thanks{Zhonghao Yang and Daojing He are with the Software Engineering Institute, East China Normal University, 200062, Shanghai, P.R. China, and also with the School of Computer Science and Technology, Harbin
Institute of Technology, 518055, Shenzhen, P.R. China.}
\thanks{Cheng Luo is with the School of Computer Science and Technology, Harbin
Institute of Technology, 518055, Shenzhen, P.R. China.}
\thanks{Yiming Li is with the Nanyang Technological University, 639798, Singapore.}
\thanks{Yu Li is with College of Integrated Circuits, Zhejiang University, Hangzhou, 310000, Zhejiang, China (e-mail: yu.li.sallylee@gmail.com).}

\thanks{Yu Li and Daojing He are the corresponding author of this article.}
}

\markboth{IEEE TRANSACTIONS ON INFORMATION FORENSICS AND SECURITY, VOL. 20, 2025}{YANG \MakeLowercase{\textit{et al.}}: ARCGEN: GENERALIZING NEURAL BACKDOOR DETECTION ACROSS DIVERSE ARCHITECTURES}

\IEEEpubid{1556-6021 © 2025 IEEE}

\maketitle

\begin{abstract}
    Backdoor attacks pose a significant threat to the security and reliability of deep learning models. 
    To mitigate such attacks, one promising approach is to learn to extract features from the target model and use these features for backdoor detection.
    However, we discover that existing learning-based neural backdoor detection methods do not generalize well to new architectures not seen during the learning phase.
    In this paper, we analyze the root cause of this issue and propose a novel black-box neural backdoor detection method called \tool. Our method aims to obtain architecture-invariant model features, i.e., \textit{aligned features}, for effective backdoor detection. 
    Specifically, in contrast to existing methods directly using model outputs as model features, we introduce an additional alignment layer in the feature extraction function to further process these features. 
    This reduces the direct influence of architecture information on the features. 
    Then, we design two alignment losses to train the feature extraction function. These losses explicitly require that features from models with similar backdoor behaviors but different architectures are aligned at both the distribution and sample levels. 
    With these techniques, our method demonstrates up to 42.5\%  improvements in detection performance (\eg, AUC) on unseen model architectures.  This is based on 
     a large-scale evaluation involving 16,896 models trained on diverse datasets, subjected to various backdoor attacks, and utilizing different model architectures. 
    Our code is available at https://github.com/SeRAlab/ArcGen.
\end{abstract}

\begin{IEEEkeywords}
Backdoor detection, Security, Deep learning.
\end{IEEEkeywords}

\section{Introduction}
\label{sec:intro}
\IEEEPARstart{D}{eep} Learning (DL) models have shown remarkable performance in various domains \cite{hu2023_uniad, mirskyvulchecker}; however, they are vulnerable to backdoor attacks. The attacker often inserts hidden backdoors into the model during the training phase, causing the model to misclassify inputs with specific triggers while behaving normally on clean inputs. The backdoor threat arises when the training process involves untrustworthy third-party datasets, platforms, or deploying models directly from such sources, jeopardizing the overall security of the models \cite{li2022backdoor, Feng_2022_CVPR}. Hence, it becomes imperative to detect the existence of such backdoors before model deployment.

There are two mainstream methods in the literature for detecting neural backdoors before deployment: training data analysis \cite{20190506448912, pmlr-v139-hayase21a, Al_Kader_Hammoud_2023_CVPR} and model analysis \cite{Wang2019, Chen2019, Liu2019, Huang2019, Kolouri2020, Xu2021a, Huang2020, zheng2021topological, sun2023smoothinv}. 
The former focuses on identifying potential poisoning of the model training data, while the latter examines model behaviors for backdoor detection.
Model analysis does not require access to private training data and is often considered more practical. 
One representative approach for model analysis is trigger reverse engineering \cite{Wang2019, Chen2019, Liu2019, sun2023smoothinv}, wherein researchers investigate the existence of backdoor triggers leading the model to target classes.
However, these methods often require white-box access to models, such as model gradients, for trigger optimization -- a requirement that is often challenging to fulfill in practical scenarios. 
For instance, for a cloud-based model service, the only available information from the model is the outputs.

A promising method to address the above issue is through learning a query-based feature extraction function, where model behaviors are reflected by features extracted by specific queries \cite{Xu2021a, Kolouri2020}. 
Given a bunch of benign and backdoored models, existing methods often learn to optimize this function (\eg, queries) such that the extracted features are distinguishable between benign and backdoored models.
Once queries are obtained, they are fed into any target model to obtain its features for backdoor detection. 
During the detection phase, model internals are not required and are thus kept confidential. 

\begin{table}[tb]
\centering
    \caption{Detection results for models with different architectures (on GTSRB dataset \cite{stallkamp2012mangtsrb}). 
    {\colorbox[HTML]{9AFF99}{Green}} and {\colorbox[HTML]{FFCCC9}{red}} represent seen and unseen architectures, respectively.
    }
\label{tab:intro}
\renewcommand{\arraystretch}{0.8}
\resizebox{\columnwidth}{!}{%
\begin{tabular}{ccccc}
\toprule
 & \multicolumn{4}{c}{\textbf{Detection Performance (Area Under the ROC Curve, AUC)}} \\ \cmidrule{2-5} 
\multirow{-3}{*}{\textbf{Methods}} & Simple CNN & SENet18\cite{hu2018senet} & ResNet18\cite{he2016deep} & MobileNetV2\cite{sandler2018mobilenetv2} \\ \midrule
 & {\cellcolor[HTML]{9AFF99} 0.9763} & {\cellcolor[HTML]{FFCCC9}0.5440} & {\cellcolor[HTML]{FFCCC9}0.5528} & {\cellcolor[HTML]{FFCCC9}0.4664} \\
\multirow{-2}{*}{\textbf{\begin{tabular}[c]{@{}c@{}}MNTD\\\cite{Xu2021a}\end{tabular}}} & {\cellcolor[HTML]{9AFF99} 0.9671} & {\cellcolor[HTML]{9AFF99} 0.8615} & {\cellcolor[HTML]{FFCCC9}0.7399} & {\cellcolor[HTML]{FFCCC9}0.5294} \\
 & {\cellcolor[HTML]{9AFF99} 0.9448} & {\cellcolor[HTML]{FFCCC9}0.4630} & {\cellcolor[HTML]{FFCCC9}0.5631} & {\cellcolor[HTML]{FFCCC9}0.5329} \\
\multirow{-2}{*}{\textbf{\begin{tabular}[c]{@{}c@{}}ULPs\\\cite{Kolouri2020}\end{tabular}}} & {\cellcolor[HTML]{9AFF99} 0.9197} & {\cellcolor[HTML]{9AFF99} 0.7005} & {\cellcolor[HTML]{FFCCC9}0.5897} & {\cellcolor[HTML]{FFCCC9}0.5357} \\ \bottomrule
\end{tabular}%
}
\end{table}
\begin{figure}[tb]
\centering
\includegraphics[width=\linewidth]{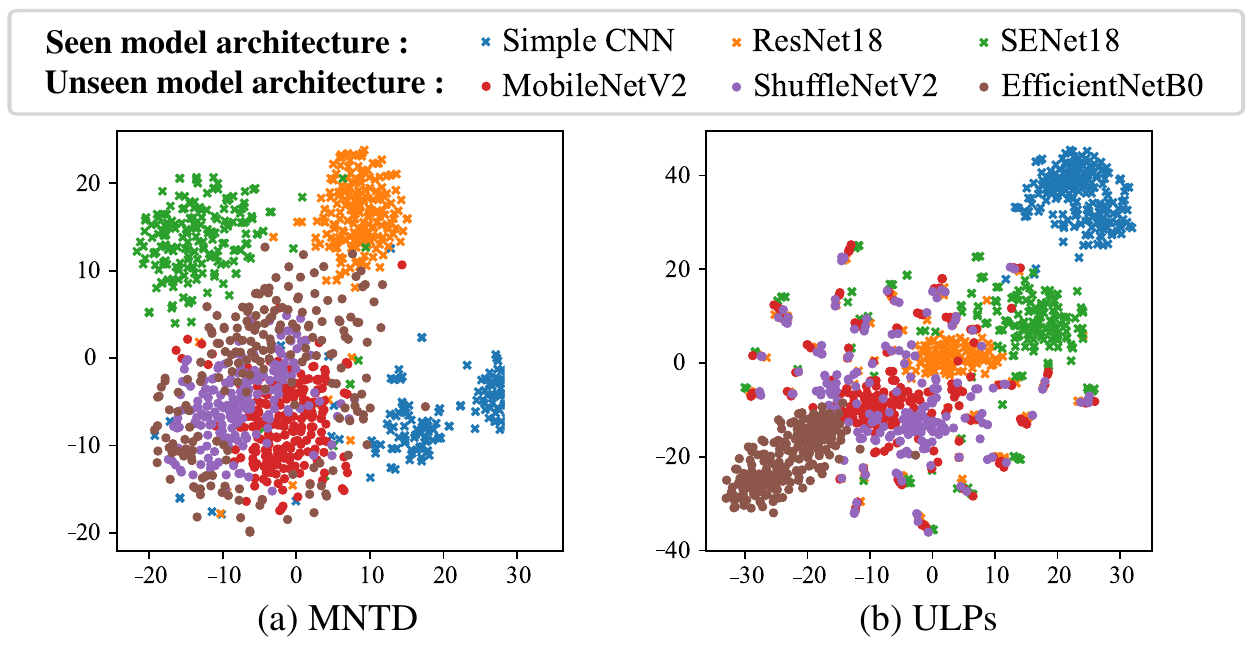}
\caption{The t-SNE visualizations of features extracted by (a) MNTD and (b) ULPs for both seen and unseen model architectures on the GTSRB dataset. 
}
\label{fig:feature_dis}
\end{figure}

\IEEEpubidadjcol
However, we observe that \textit{the learned feature extraction function does not generalize well to new architectures that were not seen during the training phase}. 
To illustrate, we present an example in Table~\ref{tab:intro}. 
The detection performance of representative methods on BadNets attack \cite{Gu2019}  shows higher performance on architectures used for training the feature extraction function (defined as \textit{seen architectures}) while exhibiting significant degradation for new architectures, \ie, \textit{unseen architectures}.
We further analyze the underlying reasons for this performance degradation. The results in Figure~\ref{fig:feature_dis} show that feature distributions extracted from different model architectures vary significantly when using the same queries, particularly for unseen architectures.
These distribution differences pose a significant challenge for backdoor detection, due to the difficulty in handling unknown feature distributions from unseen model architectures. 
Given that the model architecture is often unknown in advance in a black-box setting, it is necessary to enhance the generalization capability of the detection methods.

To address the above challenge, we present \tool, a novel black-box neural backdoor detection method that effectively generalizes to unseen model architectures. 
Our insight is to minimize the impact of model architectures on the extracted features, \ie, learn a feature extraction function so that models with similar backdoor behaviors have aligned features.
To achieve this, unlike previous methods that directly use the model outputs for specific queries as model features, \tool incorporates extra alignment layers to further process the model outputs to compensate for the architecture variations and enable effective extraction of aligned features. 
The inclusion of this alignment layer is crucial; without it, the model features would be directly influenced by the model architecture, making it challenging to eliminate the architectural effects.

To learn the feature extraction function, including both the queries and alignment layers, \tool introduces two losses: the distribution-level and the sample-level alignment loss.
The former encourages features from models with similar backdoor behaviors but different architectures to become indistinguishable at the distribution level. 
However, quantifying distribution differences in the high-dimensional feature space poses a challenge. 
Drawing inspiration from domain generalization methods, we adopt an architecture discriminator to address this -- the discriminator is trained to differentiate if the extracted features belong to specific model architectures. 
This discriminator and the feature extraction function undergo an adversarial training process so that the extracted features can achieve distribution-level alignment.
The latter, sample-level alignment loss, is designed to enforce precise feature alignment for the same backdoor on different architectures. 
The above techniques make features less susceptible to architectural biases, thereby focusing on discerning backdoor presence. 

Our contributions are summarized as follows:

\begin{itemize}

\item To our knowledge, \tool is the first neural backdoor detection method that addresses the impact of model architectures. Through aligned feature extraction, it shows remarkable generalization to various model structures.

\item We introduce alignment layers into the feature extraction process, enabling the accommodation of architecture variations within these layers. Moreover, we propose distribution-level and sample-level alignment techniques to effectively train the feature extraction function, enabling the extraction of architecture-invariant features. 

\item We further improve \tool's generalization ability by proposing architecture randomization, a simple method that turned out to be effective in enhancing our detection performance.

\end{itemize}

We conduct extensive experiments on popular datasets such as MNIST \cite{yann1995mnist}, CIFAR-10 \cite{krizhevsky2009cifar}, GTSRB \cite{stallkamp2012mangtsrb}, and ImageNet \cite{ILSVRC15}.
Detection results on 16,896 models with various architectures and attack types demonstrate that \tool outperforms existing methods, particularly in generalization ability. 
We will release our models to foster further research in this area.

\section{Related Works}
\subsection{Neural Backdoor}
\label{sec:nbbackground}
A neural backdoor refers to a hidden behavior injected into a deep learning model. The model operates normally with clean inputs but displays abnormal behavior when triggered by specific patterns.
The primary method for inserting a neural backdoor is the poisoning-based backdoor attack \cite{Gu2019, Chen2017, liu2018trojaning, Barni2019, nguyen2021wanet, wang2022bppattack, bagdasaryan2021blind, nguyen2020input}. In this attack, attackers poison training samples by inserting triggers into the images. 
For instance, BadNets \cite{Gu2019} uses a patch to serve as backdoor triggers and adds the patch to the training images. 
In the Blended attack \cite{Chen2017}, the trigger pattern is blended into the benign image with a transparency ratio for stealthy poisoning. 
The SIG attack \cite{Barni2019} introduces sinusoidal stripes onto the clean image. WaNet \cite{nguyen2021wanet} utilizes geometric transformations on the clean image to induce subtle warping effects as triggers. BppAttack \cite{wang2022bppattack} takes color bit depth change as the imperceptible trigger, generated by image quantization for the bit-per-pixel (BPP) reduction and image dithering for artifact removal. 

\subsection{The Learning-based Backdoor Detection}
\label{sec:learning_based}
Backdoor detection aims to determine whether a DL model contains a backdoor. Existing backdoor detection methods primarily focus on poisoning-based backdoor attacks \cite{li2022backdoor}, and they can be classified into three types: input-based detection, dataset-based detection, and model-based detection \cite{Xu2021a}. Input-based detection checks each input sample for the existence of backdoor triggers \cite{Gao2019, chou2020sentinet, wang2023mm}, which is useful after model deployment. To detect backdoors before model deployment, dataset-based detection tries to identify poisoned samples in the training dataset \cite{pmlr-v139-hayase21a, 20190506448912}, while model-based detection analyzes model behaviors to decide if a model is backdoored \cite{Wang2019, Chen2019, Liu2019}. We focus on model-based detection methods since they do not require access to the training dataset and avoid delays associated with online detection.

Model-based detection can be classified into white-box \cite{Wang2019, Chen2019, Liu2019, sun2023smoothinv, liu2022complex, Huang2019, zheng2021topological} and black-box methods \cite{dong2021black, guo2022aeva, Huang2020, wang2023mm, Xu2021a, Kolouri2020}. \change{Black-box methods refer to detection techniques that do not rely on access to the internal details of the target model, such as its architecture, parameters, or gradients. These approaches} are more practical than white-box methods as the model internals (e.g., parameters) are often considered intellectual properties \cite{oliynyk2023know, dong2021black}. 
For black-box detection, some studies design explicit rules to identify backdoor behaviors. \cite{dong2021black} and \cite{ guo2022aeva} focus on reverse engineering triggers for the target model. \cite{wang2023mm} analyzes the statistics of the model output and propose an anomaly detector for backdoor detection. \cite{Huang2020} suggests extracting a one-pixel signature from the target model for detection.
Simultaneously, another line of research focuses on automatically learning to extract model features for backdoor analysis \cite{Xu2021a, Kolouri2020}. This learning-based approach avoids human biases, showing promise in detecting various types of attacks.

\begin{figure}[tb]
\centering
\includegraphics[scale=0.18]{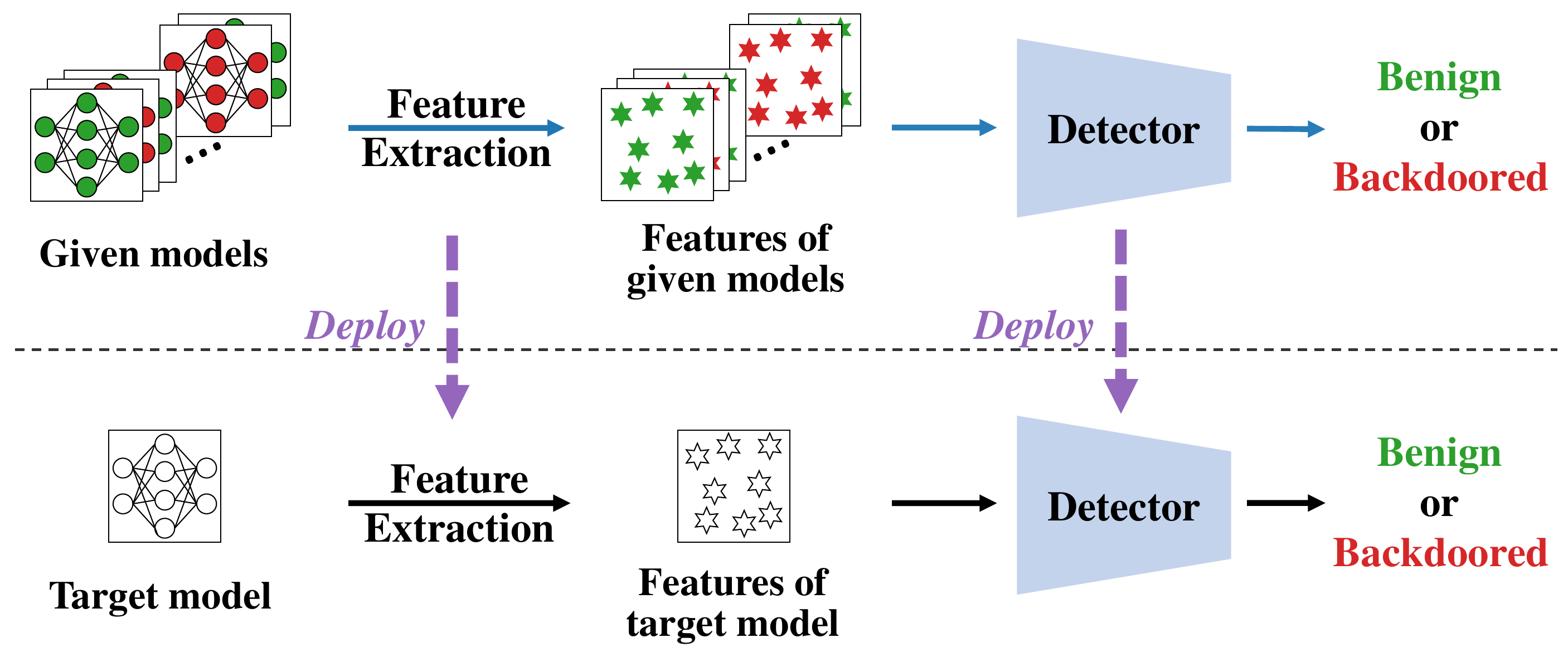}
\caption{The workflow of the learning-based detection.}
\label{fig:mfbdetection}
\end{figure}

The workflow of the learning-based approach is illustrated in Figure~\ref{fig:mfbdetection}. Given a collection of models, including both backdoored and benign ones, the approach optimizes a function to extract backdoor-related features and simultaneously trains a detector to identify whether a model is backdoored using these features.
To extract model features with only black-box access, they utilize the model output for some queries as model features. Once the feature extraction function and the detector are developed, they are deployed to assess a target model. Specifically, the optimized queries are input into the target model to extract backdoor-related features, which are then used by the detector for identification.

Two examples of this learning-based approach are Meta Neural Trojan Detection (MNTD) \cite{Xu2021a} and Universal Litmus Patterns (ULPs) \cite{Kolouri2020}. MNTD uses shadow models to simulate benign and backdoor features, while ULPs train detectors using real-world models. Both methods enable fast detection, requiring only a single forward pass after deployment.

However, despite their effectiveness, we observe that existing learning-based detection methods perform poorly on models with different architectures from the ones used to train the feature extraction function. Thus, we focus on improving the generalization ability of this type of method.

\subsection{Domain Generalization}
Domain generalization addresses the challenge of performance decline in DL models when they encounter previously unseen domain data after being trained on specific domains \cite{wang2022generalizing}. 
Our approach is inspired by two prominent domain generalization techniques.

The first technique is domain randomization, which counters overfitting to specific domains by augmenting training data with randomized domain information \cite{domainrandom2018Peng, domainrandom2018Tremblay, domainrandom2019Khirodkar, domainrandom2019Yue}. This approach enriches the source domain, mitigating the risk of overfitting.

The second technique is adversarial training, which focuses on cultivating domain-invariant features, as demonstrated by the Domain-Adversarial Neural Network (DANN) \cite{ganin2015unsupervised}. While initially devised for domain adaptation, DANN's principles can be adapted for domain generalization. DANN utilizes a generator and a discriminator, where the discriminator (domain classifier) distinguishes between different domains, and the generator (feature extractor) is trained to deceive the discriminator, facilitating the learning of domain-invariant representations. This adversarial training process enhances the extraction of features invariant to domain shifts. Moreover, several other adversarial training methods for domain generalization have been proposed \cite{li2018deep, rahman2020correlation, sicilia2023domain}. For further details, please refer to their respective papers if interested.

\section{Threat Model}
\textbf{The Attacker.}
Attackers aim to embed backdoors into deep learning (DL) models. These compromised models perform normally with benign inputs but intentionally misclassify when exposed to adversary-defined inputs. We assume strong adversaries with complete knowledge of the models and full control over the training process, including the ability to manipulate training data and fine-tune model parameters.

\textbf{The Defender.}
The defender's goal is to make a binary decision regarding the presence of a backdoor in the target model. They operate under limited knowledge and capabilities, characterized by:
\begin{enumerate}
\item{No prior attack knowledge. We do not assume any specific attack methods, such as trigger mechanisms.} 
\item{Black-box model access. Respecting the model provider's intellectual property rights, we only assume black-box access to the model, allowing us to provide inputs and obtain predicted probabilities for different classes.}
\item{Assumed knowledge of training data distribution. We follow \cite{Xu2021a, Kolouri2020} and assume that the defender cannot access the model's specific training data but is aware of its overall distribution.}
\end{enumerate}

Armed with this information, the defender can select appropriate open-source model architectures to train given models (e.g., benign and proxy backdoor models).

\section{Problem Formulation}
\label{sec:problem_formulation}
In a task of neural backdoor detection, let $\mathcal{M}$ and $\mathcal{B}$ represent the input and output space, respectively. Then $m \in \mathcal{M}$ denotes a model, and $b \in \mathcal{B} \subset \mathbb{R}$ is a binary label indicating whether the model is compromised by backdoor attacks. 
An example set $\mathcal{A} = \{(m_i, b_i)\}_{i=1}^n \sim P_{MB}$ for a model architecture is a collection of data sampled from the joint distribution $P_{MB}$, where $P_{MB}$ denotes the joint distribution of models $M$ and labels $B$.

Assuming there are $N$ given example sets for different model architectures, denoted as $\mathcal{A}_{given} = \{\mathcal{A}^1, \mathcal{A}^2, \ldots, \mathcal{A}^N \}$. Each example set $\mathcal{A}^i = \{(m_j^i, b_j^i)\}_{j=1}^{n_i}$ contains models with both benign and backdoored features. Because of the different architectures, the joint distributions between each pair of example sets are different: $P_{MB}^i \neq P_{MB}^j \text{ for } 1 \leq i \neq j \leq N$. The objective is to learn a detection function $g: \mathcal{M} \rightarrow \mathcal{B}$ from the $N$ given example sets $\mathcal{A}_{given}$ to minimize the prediction error on an unseen target example set $\mathcal{A}_{target}$. The target example set $\mathcal{A}_{target}$ contains models with architectures different from those in $\mathcal{A}_{given}$ and cannot be accessed during training. Hence, its joint distribution $P_{MB}^{target}$ is different from the distribution $P_{MB}^i$ for each set of training examples. The optimization objective can be written as:
\begin{equation}
\min_{g} \; \mathbb{E}_{(m,b) \in \mathcal{A}_{target}} \left[ \ell_{t}(g(m), b) \right],
\end{equation}
where $\mathbb{E}$ is the expectation and $\ell_{t}(\cdot, \cdot)$ is the loss function measuring the difference between the prediction by the function $g$ and the true label $b$ on the unseen target example set.

\section{Methodology}
This section presents the details of \tool, a learning-based neural backdoor detection method that demonstrates high generalization ability across diverse architectures.

\begin{figure*}[tb]
\centering
\includegraphics[width=0.8\linewidth, scale=0.33]{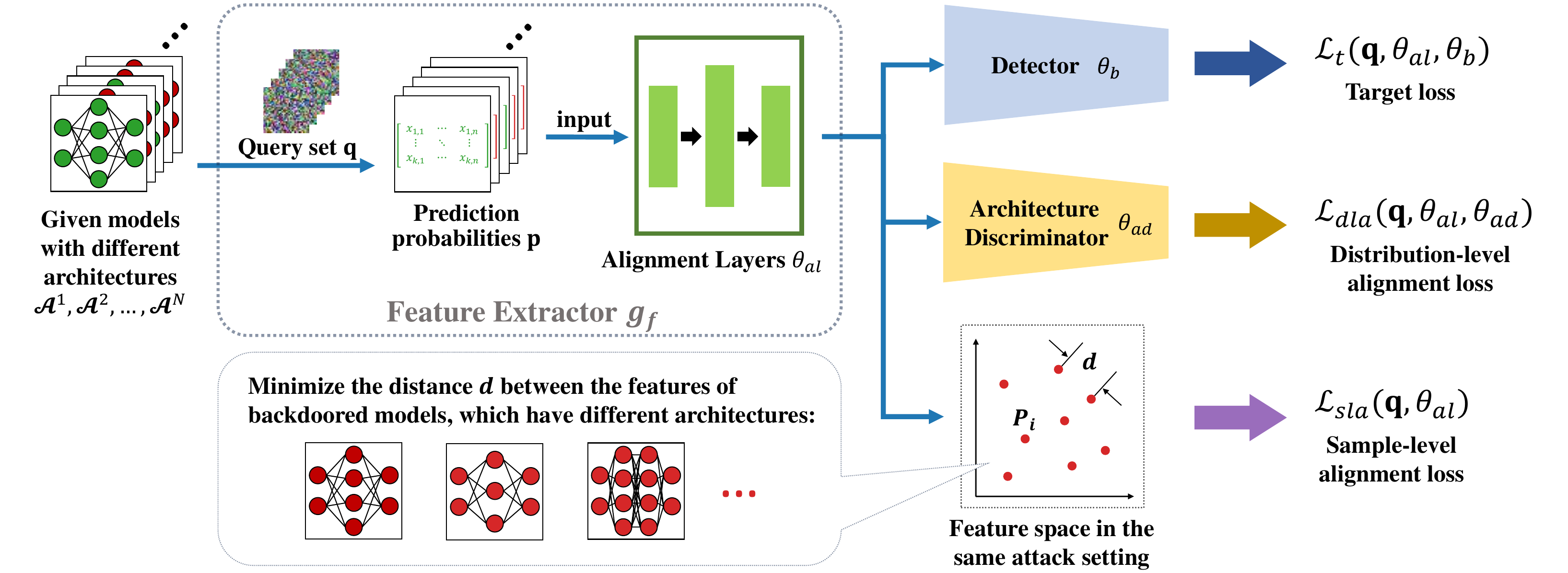}
\caption{
An overview of the proposed \tool. 
The system comprises two main components: the feature extractor and the detector. The feature extractor is designed to extract aligned features from various model architectures, while the detector identifies backdoors based on these extracted features.
To train the feature extractor effectively, we incorporate two additional losses—distribution-level alignment and sample-level alignment losses. These losses ensure that feature distributions remain indistinguishable across different architectures. The distribution-level alignment loss is achieved through adversarial learning with an architecture discriminator. Concurrently, the sample-level alignment is applied to a group of backdoored models that share the same attack settings but differ in architecture. This promotes the concentration of features from these models within the feature space, enhancing the extractor's ability to generalize across diverse architectures.
}
\label{fig:overview_fig}
\end{figure*}

\subsection{Architecture of \tool}
\label{sec:allayer}

\tool{} operates in two stages: training and detection. In the training stage, the defender utilizes open-source model architectures to train both benign and proxy backdoored models, leveraging knowledge of the training data distribution. In the detection stage, the defender, lacking prior knowledge of the target model's architecture or attack parameters, inputs a query set into the target model and employs the detector to process the model outputs for detection. The architecture of \tool{} is illustrated in Figure~\ref{fig:overview_fig}.

To achieve our goal of finding a mapping function $g: \mathcal{M} \rightarrow \mathcal{B}$ that generalizes across model architectures, we decompose $g$ into a feature extractor and a detector: $g(\cdot) = g_b(g_f(\cdot))$. The feature extractor $g_f$ learns to map examples to a $d$-dimensional feature space $\mathcal{E} \subset \mathbb{R}^d$, while the detector $g_b$ learns to map features to the output space $\mathcal{B}$. This decomposition facilitates control over the feature space, enabling aligned feature learning.
To determine if a given model contains backdoors, a set of queries $\mathbf{q}$ is input into the model to obtain its output probabilities for different classes, denoted as $\mathbf{p} = m(\mathbf{q})$. These probabilities are then processed through alignment layers ${al}(\cdot;\theta_{al})$ to extract backdoor-related features $\mathbf{e}$. The entire feature extraction process is represented as $\mathbf{e}={g_f}(m;\mathbf{q}, \theta_{al})$. The architecture-invariant features $\mathbf{e}$ are fed into the detector, which outputs a scalar representing the likelihood of the model being benign or backdoored.

\textbf{Alignment Layers (ALs).} 
\change{Since $\mathbf{p}$ is a direct output of the model $m$, it is strongly associated with the model architecture and may embed architecture-specific biases that cannot be fully mitigated through optimization over the input queries $\mathbf{q}$. These biases stem from the differing inductive priors and learned representations of different architectures, leading to divergent output distributions even when models are trained on the same dataset.}

\change{Empirically, we observe that the output distributions differ for almost every pair of distinct architectures, this discrepancy can also be approximated by significant differences in test accuracy across architectures, as shown in Table~\ref{tab:arc_acc}. The results demonstrate that, for almost every pair of architectures' accuracies on the same test dataset differ noticeably. This accuracy gap serves as a practical reflection of the underlying divergence in their architecture-level output distributions. Furthermore, Figure~\ref{fig:feature_dis} (in Section~\ref{sec:intro}) provides additional evidence by visualizing architecture-level output distributions. It shows that models of different architectures tend to produce outputs that occupy distinct regions in high-dimensional space. Each color in the figure represents the distribution of one specific architecture, highlighting the separation among them. \textbf{These discrepancies result in distribution shifts that severely degrade detection performance on unseen architectures, as the detector may encounter out-of-distribution features during inference.}}

To mitigate this issue, we introduce a learnable transformation module termed the \textbf{alignment layers} ${al}(\cdot)$, applied directly to the predicted probability vector $\mathbf{p}$. The alignment layers aim to project outputs from different architectures into a shared feature space, yielding architecture-invariant features $\mathbf{e} = al(\mathbf{p})$. This alignment is theoretically supported by the universal approximation theorem \cite{cybenko1989approximation}, which states that a sufficiently expressive neural network can approximate any measurable function. Accordingly, our alignment layers are trained to minimize the distribution divergence between aligned outputs while preserving the semantics necessary for backdoor detection.

In practice, the alignment layers are optimized jointly with the rest of the feature extractor under a multi-objective framework combining a distribution-level alignment loss $\mathcal{L}_{dla}$ and a sample-level alignment loss $\mathcal{L}_{sla}$, along with the task loss $\mathcal{L}_t$. This joint training enables the alignment layers to effectively balance architecture invariance and task discriminability. Evaluation in Section~\ref{sec:ablation_exp} demonstrates that these layers are critical for achieving robust generalization to unseen architectures.

\begin{table*}[t]
\caption{\change{Test accuracies of benign models across architectures on the GTSRB dataset.}}
\label{tab:arc_acc}
\begin{tabular}{@{}cccccccccc@{}}
\toprule
\textbf{Dataset} & \textbf{Simple CNN} & \textbf{MobileNetV2} & \textbf{ResNet18} & \textbf{EfficientNetB0} & \textbf{SENet18} & \textbf{ShuffleNetV2} & \textbf{ViT-B/16} & \textbf{ViT-B/32} & \textbf{ViT-L/32} \\ \midrule
\textbf{GTSRB} & 0.9227 & 0.9683 & 0.9697 & 0.9415 & 0.9669 & 0.9436 & - & - & - \\
\textbf{MNIST} & 0.9898 & 0.8562 & 0.9937 & 0.9895 & 0.9939 & 0.9928 & - & - & - \\
\textbf{CIFAR-10} & 0.8588 & 0.9155 & 0.9223 & 0.8539 & 0.9206 & 0.9040 & - & - & - \\
\textbf{ImageNet} & - & 0.9002 & 0.8149 & 0.7811 & 0.8372 & 0.7885 & 0.9644 & 0.9547 & 0.9615 \\ \bottomrule
\end{tabular}%
\end{table*}

During training, the feature extractor $g_f$ is optimized with a distribution-level alignment loss $\mathcal{L}_{dla}$ to ensure that the features are indistinguishable across architectures, thus focusing more on extracting backdoor-related features. This is achieved through an adversarial learning process with an architecture discriminator $g_{ad}$. We also propose the sample-level alignment loss $\mathcal{L}_{sla}$ to further align model features by encouraging the features of proxy backdoored models with the same attack settings to be closer to each other in the feature space. These losses, along with the target loss $\mathcal{L}_{t}$, are employed jointly to optimize the feature extractor $g_f$, detector $g_b$, and architecture discriminator $g_{ad}$. Architecture randomization is incorporated to enhance the generalization ability of \tool{}.

\subsection{Theoretical Justification of the Alignment Layer}

\subsection{Target Loss}
During training, we optimize the detection function ${g}$ on a given example set $\mathcal{A}_{given}$ containing models with different architectures. The optimization objective is given by:
\begin{equation}
\begin{split}
\min_{\mathbf{q},\theta_{al},\theta_{b}} \; & {\mathcal{L}_{t}(\mathbf{q},\theta_{al},\theta_{b})} = \\ & \frac{1}{\sum_{i=1}^{N} n_i} \sum_{i=1}^{N}\sum_{j=1}^{n_i} \ell_t({g}(m_j^i;\mathbf{q},\theta_{al},\theta_{b}), b_j^i),
\label{equ:oldgopt}
\end{split}
\end{equation}
where $\ell_t(\cdot, \cdot)$ is the target loss function used to measure the discrepancy between the prediction by $g$ and the label $b$, which indicates whether the model $m$ is benign or proxy backdoored.

\subsection{Distribution-level Alignment Loss}
\label{distribution_loss}
Due to the challenge of distinguishing distributions in high-dimensional spaces, we employ an adversarial learning process to ensure the extraction of architecture-invariant features.

During training, feature extractor $g_f$ engages in competition with the architecture discriminator $g_{ad}$. The primary objective of the architecture discriminator is to identify the model architecture based on the features extracted by the feature extractor. By introducing this adversarial component, the feature extractor is incentivized to learn features that make it difficult for the architecture discriminator to distinguish between different architectures.

Specifically, for the given training sample set $\mathcal{A}_{given}$, the architecture discriminator $g_{ad}$ learns a function $g_{ad}: \mathcal{E} \rightarrow \mathbb{R}^N$, where $\mathbb{R}^N$ represents the probability distribution over the $N$ given architectures. For any model $m_j^i$ in $\mathcal{A}_{given}$, we extract features $\mathbf{e}_j^i$ using $g_f$ and input them to $g_{ad}$ to obtain the probability $\mathbf{d}_j^i$ that model $m_j^i$ is from each set $\mathcal{A}_{given}^i$ in $\mathcal{A}_{given}$. The architecture discrimination process for model $m_j^i$ is represented as:
\begin{equation}
{\mathbf{d}_j^i}=g_{ad}(g_f(m_j^i;\mathbf{q},\theta_{al});\theta_{ad}).
\end{equation}

To enable $g_{ad}$ to effectively identify the model architectures from the extracted features, it is optimized with the objective:
\begin{equation}
    \begin{split}
    \min_{\theta_{ad}} \; &{\mathcal{L}_{ad}(\theta_{ad})} = \\ & \frac{1}{\sum_{i=1}^{N} n_i} \sum_{i=1}^{N}\sum_{j=1}^{n_i} \ell_{ad}(g_{ad}(g_f(m_j^i;\mathbf{q},\theta_{al});\theta_{ad}), i) ,
    \end{split}
\end{equation}
where $\mathcal{L}_{ad}(\theta_{ad})$ is the discriminator loss, and $\ell_{ad}(\cdot, \cdot)$ is the discriminator loss function used to measure the discrepancy between the architecture prediction by the function $g_{ad}$ and the real architecture. As there are multiple given architectures for classification, we utilize the standard cross-entropy loss.

The architecture discriminator aims to detect model architectures from the features extracted by $g_f$, while the feature extractor aims to ensure that the extracted features are architecture-invariant. To achieve this, we design the distribution-level alignment loss to force the features extracted by the feature extractor to be indistinguishable by the architecture discriminator.  Specifically, this loss is added to the optimization process of $g_f$:
\begin{equation}
\begin{split}
& \min_{\mathbf{q},\theta_{al}} \; {\mathcal{L}_{dla}(\mathbf{q},\theta_{al})} = \\ & - \frac{1}{\sum_{i=1}^{N} n_i} \sum_{i=1}^{N}\sum_{j=1}^{n_i} \ell_{ad}(g_{ad}(g_f(m_j^i;\mathbf{q},\theta_{al});\theta_{ad}), i) ,
\end{split}
\end{equation}

During the learning procedure, the parameters $\mathbf{q}$ and $\theta_{al}$ of the feature extractor are optimized by maximizing the discriminator loss, while the parameters $\theta_{ad}$ of the architecture discriminator are optimized with the opposite objective.

\subsection{Sample-level Alignment Loss}
\label{sample_loss}
To further reduce the model architecture information in the extracted features $\mathbf{e}$, we employ sample-level alignment loss. 
We consider that models with the same attack settings (e.g., the trigger pattern and target label) but different architectures should exhibit consistent behavior for the same query set. 
Therefore, we introduce the sample-level alignment loss to encourage models with different architectures but the same attack settings to have the same features.
Specifically, from the given example set $\mathcal{A}_{given}$, for all proxy backdoored examples, we find examples with the same attack settings as a set. Assuming we have $w$ different sets $\{\mathcal{T}^i \mid i=1, \ldots, w\}$, each set $\mathcal{T}^i$ contains $n^i$ models with the same attack settings. We then add the following optimization objective to the optimization process of ${g_f}$:
\begin{equation}
    \begin{split}
    \min_{\mathbf{q},\theta_{al}} \; & \mathcal{L}_{sla}(\mathbf{q},\theta_{al}) = \\ & \sum_{i=1}^{w} (\frac{1}{n^i(n^i-1)} \sum_{\substack{m_k, m_l \in \mathcal{T}^i \\ k \neq l}} D_C(g_f(m_k), g_f(m_l))),
    \end{split}
\end{equation}
where $D_C(\cdot, \cdot)$ represents the cosine distance between the extracted model features.

\subsection{Joint Optimization}
\label{sec:optimazation}

The optimization of \tool involves two main steps: detection optimization with distribution-level and sample-level alignment. During each epoch, all examples in $\mathcal{A}_{given}$ are processed, with $\{\mathcal{T}^i \mid i=1, \ldots, w\}$ traversed every $\tau_{epoch}$ epochs for sample-level alignment. In the detection optimization with distribution-level alignment, the feature extractor $g_f$ is optimized to extract features that are non-discriminative regarding model architectures but can still distinguish between benign and backdoored models. This is achieved by optimizing $\mathcal{L}_{t}(\mathbf{q},\theta_{al},\theta_{b})$ and $\mathcal{L}_{dla}(\mathbf{q},\theta_{al})$ simultaneously. To optimize $g_f$ and the architectural discriminator $g_{ad}$ concurrently, a gradient reversal layer (GRL) \cite{ganin2015unsupervised} is inserted between them. During forward propagation, the GRL acts as an identity transformation, but during backpropagation, it reverses the gradient sign before passing it to the preceding layer. 

The learning process updates the parameters $\mathbf{q}$, $\theta_{al}$, and $\theta_{b}$ of $g$, as well as the parameters $\theta_{ad}$ of the architecture discriminator following the equations below:
\begin{equation}
\mathbf{q} \leftarrow \mathbf{q} - \alpha_{lr} (\nabla \mathcal{L}_{t}(\mathbf{q},\theta_{al},\theta_{b}) + {\lambda}_{dla} \nabla \mathcal{L}_{dla}(\mathbf{q},\theta_{al})),
\end{equation}
\begin{equation}
\theta_{al} \leftarrow \theta_{al} - \alpha_{lr} (\nabla \mathcal{L}_{t}(\mathbf{q},\theta_{al},\theta_{b}) + {\lambda}_{dla} \nabla \mathcal{L}_{dla}(\mathbf{q},\theta_{al})),
\end{equation}
\begin{equation}
\theta_{b} \leftarrow \theta_{b} - \alpha_{lr} \nabla \mathcal{L}_{t}(\mathbf{q},\theta_{al},\theta_{b}),
\end{equation}
\begin{equation}
\theta_{ad} \leftarrow \theta_{ad} - \alpha_{lr} {\lambda}_{dla} \nabla \mathcal{L}_{ad}(\mathbf{q},\theta_{al}),
\end{equation}
where $\alpha_{lr}$ is the learning rate and $\lambda_{dla}$ is a hyper-parameter used to control the trade-off between the two losses during training.

During the sample-level alignment process, $\mathcal{L}_{sla}(\mathbf{q},\theta_{al})$ is used to ensure that models with different architectures but the same attack settings have features that are close in the feature space. The parameters $\mathbf{q}$ and $\theta_{al}$ of $g$ are updated as follows:
\begin{equation}
\mathbf{q} \leftarrow \mathbf{q} - \alpha_{lr} \nabla \mathcal{L}_{sla}(\mathbf{q},\theta_{al}),
\end{equation}
\begin{equation}
\theta_{al} \leftarrow \theta_{al} - \alpha_{lr} \nabla \mathcal{L}_{sla}(\mathbf{q},\theta_{al}).
\end{equation}

\newcommand\mycommfont[1]{\scriptsize\ttfamily\textcolor{brown}{#1}}
\SetCommentSty{mycommfont}

\begin{algorithm}[t]
    \caption{Optimization of \tool}
    \label{alg:alg1}
    \SetAlgoNlRelativeSize{-1}
    \SetNoFillComment
    \renewcommand{\baselinestretch}{0.6} 
    \scriptsize
    \KwIn{Given set $\mathcal{A}_{given}$, $\{\mathcal{T}^i \mid i=1, \ldots, w\}$, Detection function $g(\cdot;\mathbf{q},\theta_{al},\theta_{b})$, architecture discriminator $g_{ad}(\cdot;\theta_{ad})$, loss functions $\ell_t(\cdot, \cdot)$, $\ell_{ad}(\cdot, \cdot)$, $D_C(\cdot, \cdot)$, learning rate $\alpha_{lr}$, hyper-parameter $\tau_{epoch}$, $\lambda_{dla}$}
    \KwOut{Optimized detection function $g$}
    \BlankLine
    epoch $\leftarrow$ 0\;
    \While{not converged}{
        \For{each model $m_j^i$ in $\mathcal{A}_{given}$}{
            $\mathbf{p}_j^i \leftarrow m_j^i(\mathbf{q})$ \tcp*{Get model output for queries}
            $\mathbf{e}_j^i \leftarrow {al}(\mathbf{p}_j^i;\theta_{al})$ \tcp*{Extract features using ALs}
            $\mathbf{d}_j^i \leftarrow g_{ad}(\mathbf{e}_j^i;\theta_{ad})$ \tcp*{Get discriminator output}
            $t_j^i \leftarrow gb(\mathbf{e}_j^i;\theta_{b})$ \tcp*{Get detection output}
            \tcc{Calculate target loss and distribution-level alignment loss}
            $\mathcal{L}_{t}(\mathbf{q},\theta_{al},\theta_{b}) \leftarrow \ell_t(t_j^i, b_j^i)$\;
            $\mathcal{L}_{dla}(\mathbf{q},\theta_{al}) \leftarrow - \ell_{ad}(\mathbf{d}_j^i, i)$\;
            ${\mathcal{L}_{ad}(\theta_{ad})} \leftarrow \ell_{ad}(\mathbf{d}_j^i, i)$\;
            \tcc{Update parameters}
            $\mathbf{q} \leftarrow \mathbf{q} - \alpha_{lr} (\nabla \mathcal{L}_{t}(\mathbf{q},\theta_{al},\theta_{b}) + {\lambda}_{dla} \nabla \mathcal{L}_{dla}(\mathbf{q},\theta_{al}))$\;
            $\theta_{al} \leftarrow \theta_{al} - \alpha_{lr} (\nabla \mathcal{L}_{t}(\mathbf{q},\theta_{al},\theta_{b}) + {\lambda}_{dla} \nabla \mathcal{L}_{dla}(\mathbf{q},\theta_{al}))$\;
            $\theta_{b} \leftarrow \theta_{b} - \alpha_{lr} \nabla \mathcal{L}_{t}(\mathbf{q},\theta_{al},\theta_{b})$\;
            $\theta_{ad} \leftarrow \theta_{ad} - \alpha_{lr} {\lambda}_{dla} \nabla \mathcal{L}_{dla}(\mathbf{q},\theta_{al})$\;
        }
        \tcc{Sample-level alignment every $\tau_{epoch}$ epochs}
        \If{epoch \% $\tau_{epoch}$ == 0}{
            \For{each set $\mathcal{T}^i$ in $\{\mathcal{T}^i \mid i=1, \ldots, w\}$}{
                $\mathcal{L}_{sla}(\mathbf{q},\theta_{al}) \leftarrow 0$ \tcp*{Initializing sample-level alignment loss}
                \For{each model $m_k^i$ in $\mathcal{T}^i$}{
                    $\mathbf{p}_k^i \leftarrow m_k^i(\mathbf{q})$\;
                    $\mathbf{e}_k^i \leftarrow {al}(\mathbf{p}_k^i;\theta_{al})$\;
                }
                \For{each pair of models $m_k^i, m_r^i$ in $\mathcal{T}^i$}{
                    $\mathcal{L}_{sla}(\mathbf{q},\theta_{al}) \leftarrow \mathcal{L}_{sla}(\mathbf{q},\theta_{al}) + D_C(\mathbf{e}_k^i, \mathbf{e}_r^i)$ \tcp*{Calculate and accumulate sample-level alignment loss}
                }
                $\mathcal{L}_{sla}(\mathbf{q},\theta_{al}) \leftarrow \mathcal{L}_{sla}(\mathbf{q},\theta_{al}) / {len}(\mathcal{T}^i)$\;
                $\mathbf{q} \leftarrow \mathbf{q} - \alpha_{lr} \nabla \mathcal{L}_{sla}(\mathbf{q},\theta_{al})$\;
                $\theta_{al} \leftarrow \theta_{al} - \alpha_{lr} \nabla \mathcal{L}_{sla}(\mathbf{q},\theta_{al})$\;
            }
        }
        epoch $\leftarrow$ epoch + 1\;
    }
    \Return $g$
\end{algorithm}
With this optimization process, the feature extractor $g_f$ will learn architecture-invariant $\mathbf{p}$ that allows the detector $g_b$ to accurately distinguish between benign and backdoored models.
Here, we provide a comprehensive overview of the complete joint optimization steps as Algorithm~\ref{alg:alg1}.

Additionally, inspired by domain randomization in domain generalization, we propose architecture randomization to enhance \tool's generalization to unseen architectures. 
During the training, a binary mask vector $\mathbf{mask} \in \{0, 1\}^k$ is used to randomly mask a portion of the model's weights, where $k$ represents the total number of weights in the model. The mask vector $\mathbf{mask}$ controls which weights are masked and which ones are retained. We utilize the mask vector $\mathbf{mask}$ to perform element-wise multiplication with the model's weights $\theta_m$ to obtain a masked model $m'$, represented as $\theta_{m'} = \theta_{m} \odot \mathbf{mask}$, where $\odot$ denotes element-wise multiplication. The hyper-parameter $\beta_{mask}$ controls the probability of 0 appearing in $\mathbf{mask}$, thereby controlling the probability of weight masking and the extent of the effect on the model's benign accuracy and attack success rate. Through this process, we simulate random transformations on the model's architecture, enhancing the detection method's ability to generalize to unseen architectures.

\section{Experiment}

In this section, we present a comprehensive evaluation of our proposed method. All our experiments are conducted on eight NVIDIA GeForce RTX 4090 GPUs.

\subsection{Experimental Setup}
\label{sec:expset}
\subsubsection{Datasets}
We conduct experiments on four datasets widely used in computer vision tasks: MNIST, CIFAR-10, GTSRB and ImageNet. {MNIST} \cite{yann1995mnist} consists of black and white handwritten digit images ranging from 0-9, each sized 28x28 pixels. CIFAR-10 \cite{krizhevsky2009cifar} comprises 60,000 color images, each measuring 32x32 pixels, distributed across 10 classes. {GTSRB \cite{stallkamp2012mangtsrb}} contains 43 different classes of traffic signs. ImageNet \cite{ILSVRC15} contains millions of labeled images across 1000 classes. In our experiment, we utilize a subset of ImageNet comprising 10 randomly chosen classes to mitigate the costs associated with training numerous target models.  

The defender requires sufficient data to train the given model due to the limited data volume may lead to uncertain model behavior and hinder effective feature analysis. Additionally sufficient data is needed to train the target model to ensure the models tested can exhibit good performance and are representative of those used in real-world scenarios. Therefore, for each dataset, we allocate 55\% of the training set randomly to the attacker and 45\% to the defender.

\subsubsection{Architectures}
\label{subsubsec: architecture}
We prepare an architecture pool for each dataset to evaluate \tool, and they are stated as follows:

\textit{MNIST.} For this dataset, we utilize several model architectures, including SENet18 \cite{hu2018senet}, ResNet18 \cite{he2016deep}, MobileNetV2 \cite{sandler2018mobilenetv2}, ShuffleNetV2 \cite{Ma_2018_shufflenetv2}, EfficientNetB0 \cite{tan2019aefficienet}, and a simple convolutional neural network (Simple CNN). The Simple CNN architecture comprises two convolutional layers, two max-pooling layers, two linear layers, and ReLU activation. 

\textit{CIFAR-10 and GTSRB.} 
For these two datasets, we utilize the same architectures as those used for MNIST, with the exception of the Simple CNN architecture. The Simple CNN for CIFAR-10 and GTSRB comprises four convolutional layers, two max-pooling layers, three linear layers, and ReLU activation. This optimized architecture delivers improved model accuracy for these two datasets.

\textit{ImageNet.} For this dataset, due to the low accuracy of Simple CNNs, which makes them impractical for real-world applications, we exclude them from this dataset. 
We adapt the input size to be $3 \times 224 \times 224$ for the rest of the architectures from CIFAR-10 and GTSRB. In addition, we evaluate our approach using the more advanced transformer-based models by selecting ViT-Base/16, ViT-Base/32, and ViT-Large/32 architectures with pre-trained weights \cite{dosovitskiy2020vit}.

\subsubsection{Attack Settings}
\label{subsubsec: attack_setting}
On each dataset, the attacker generates backdoored models using different architectures, including Simple CNN, SENet18, ResNet18, MobileNetV2, ShuffleNetV2, and EfficientNetB0. BadNets \cite{Gu2019} and Blended \cite{Chen2017} serve as the primary attack methods to evaluate \tool's detection performance. Furthermore, SIG \cite{Barni2019}, WaNet \cite{nguyen2021wanet}, and BppAttack \cite{wang2022bppattack} attacks are employed to assess \tool's detection performance against stealthy attacks. 
The target label for each backdoored model is uniformly chosen from the output set of each task. 
The poisoning ratio determines the percentage of poisoned samples in the training dataset and is uniformly sampled according to the specified range for each attack method.
The following are the descriptions of different attack methods:
\begin{itemize}
\item{\textit{BadNets \cite{Gu2019}.} BadNets poisons the training data by directly replacing image pixels with a specific trigger pattern. }

\item{\textit{Blended \cite{Chen2017}.} The Blended method uses a trigger pattern that covers the entire input image and is the same size as the input.}

\item{\textit{SIG \cite{Barni2019}.} The SIG method generates a trigger pattern that covers the entire input image based on sinusoidal stripes.}

\item{\textit{WaNet \cite{nguyen2021wanet}.} The WaNet method involves generating poisoned images with small distortion to preserve the content of the image while introducing a subtle warping effect as the trigger.}

\item{\textit{BppAttack \cite{wang2022bppattack}.} The BppAttack method is a novel image color quantization-based backdoor attack. It leverages image color quantization and dithering to generate high-quality attack triggers and poisoning samples.}
\end{itemize}

\subsubsection{Defense Settings}
We first introduce the model architectures the defender used to train \tool{}, then introduce the method used for proxy backdoored model generation, and finally introduce the detailed setup of the \tool{}'s component and training configuration of \tool{}.

\textbf{Seen architectures}. For the MNIST, CIFAR-10, and GTSRB datasets, the defender generates 256 proxy backdoored models and 256 benign models for each architecture: Simple CNN, SENet18, and ResNet18.
For the ImageNet dataset, Simple CNN is excluded due to its low accuracy, which makes it impractical for real-world applications. Instead, the defender generates 64 proxy backdoored models and 64 benign models for each architecture: ResNet18, MobileNetV2, and EfficientNetB0. The number of models is set to be smaller due to training time constraints.
Note that the seen architectures for all datasets are randomly selected from the architecture pool (see \ref{subsubsec: architecture} for details of the pool), and Table \ref{tab:Ediffseen} examines the impact of different combinations of the seen architectures.

\textbf{Proxy backdoored model generation.} The proxy backdoored models are generated using the method from \cite{Xu2021a}, which employs triggers with random pixel patterns, transparency, and sizes. This approach allows the detection method to demonstrate strong performance across diverse attack settings without prior knowledge of the target model's specifics (e.g., trigger size and trigger type). To facilitate effective sample-level alignment loss calculation, the defender trains these models with the same attack settings across different architectures. 

\textbf{Detailed setup of the \tool{}'s component.} \tool{} is designed in an adversarial training framework and comprises four main components: a learnable input query set, Alignment Layers (ALs), a detector, and an architecture discriminator. Table~\ref{tab:detector_arch} outlines the architecture of each module in \tool{}.

\begin{table}[t]
\centering
\caption{\change{Detailed setup of the \tool{}'s component. Here \(S_{\text{input}}\) is the query length, \(N_{\text{in}}\) is the number of queries, $d=S_{\text{output}}\times N_{in}$ is the feature dimension, and $N_{\text{arch}}$ is the number of given architectures.}}
\label{tab:detector_arch}
\resizebox{0.95\columnwidth}{!}{%
\begin{tabular}{ll}
\toprule
\textbf{Module} & \textbf{Structure} \\ \midrule
\textbf{Query set} & Learnable query set $\mathbf{q}\in\mathbb{R}^{S_{\text{input}}\times N_{\text{in}}}$, \\ 
               \addlinespace
\textbf{Alignment Layers} & FC($d,4d$) $\to$ ReLU $\to$ Dropout(0.5) \\ 
                         & FC($4d,d$) $\to$ ReLU $\to$ Dropout(0.5) \\ \addlinespace
\textbf{Detector}  & FC($d,N_h\!=\!20$) $\to$ linear output (1) \\ \addlinespace
\textbf{Architecture Discriminator}   & GRL $\to$ FC($d,512$) $\to$ ReLU $\to$ Dropout(0.5) \\ 
                         & $\quad\;\;\to$ FC($512,\;N_{\text{arch}}$) \\ \bottomrule
\end{tabular}
}
\end{table}

\change{\textbf{Training setup of \tool{}.} The key training settings for \tool{} are summarized in Table~\ref{tab:training_config}. These include optimizer choices, learning rate schedule, alignment loss hyper-parameters, and training repetitions.}

\begin{table}[t]
\centering
\caption{\change{Training setup used for training \tool{}, including optimizer configuration, alignment loss hyper-parameters, and learning schedule.}}
\label{tab:training_config}
\resizebox{0.95\columnwidth}{!}{%
\begin{tabular}{ll}
\toprule
\textbf{Parameter}                    & \textbf{Value}                          \\ \midrule
Optimizer                             & Adam (AMSGrad)                          \\
Initial learning rate                 & \(1\times10^{-2}\)                      \\
Batch size                            & 32                                       \\
Total epochs                          & 90                                       \\
Scheduler                   & StepLR: step size = 30, \(\gamma=0.6\), \\ 
                             & applied after \(\tfrac{4}{6}\text{epochs}\)  \\
Gradient clipping norm                & 1.0                                      \\
Sample-level alignment interval \(\tau_{epoch}\)   & 5                                       \\
Distribution-level loss weight \(\lambda_{dla}\)      & 1.0                                     \\
Random seed                           & 0                                        \\
Repetitions                           & 5 runs (average reported)               \\ \bottomrule
\end{tabular}
}
\end{table}

\subsubsection{Baselines}
We compared \tool{} with state-of-the-art learning-based techniques MNTD \cite{Xu2021a} and ULPs \cite{Kolouri2020}. Also, for a comprehensive assessment, we include the recently proposed non-learning-based black-box detection method, Maximum-Margin-based Backdoor Detection (MM-BD) \cite{wang2023mm}.

\subsubsection{Metrics}
We utilize benign accuracy (BA) to measure model performance on normal inputs and attack success rate (ASR) to assess the attack effectiveness of the backdoored model. For backdoor detection performance, we employ the area under the curve (AUC) metric, considering the trade-off between the true positive rate and the false positive rate.

\begin{table}[tb]
\centering
\caption{Average benign accuracy (BA) of benign models and the average BA and average attack success rate (ASR) of backdoored models for given and target models. 
The defender uses the given models to develop detectors, while the target models are used to test detection performance.
}
\label{tab:Eset}
\renewcommand{\arraystretch}{0.6}
\resizebox{0.95\columnwidth}{!}{%
\begin{tabular}{ccccc}
\toprule
\multirow{2}{*}{} & \multirow{3}{*}{\textbf{Dataset}} & \textbf{Benign model} & \multicolumn{2}{c}{\textbf{Backdoored model}} \\ \cmidrule{3-5} 
 &  & \textbf{BA} & \textbf{BA} & \textbf{ASR} \\ \midrule
\multirow{4}{*}{\textbf{Given Model}} & GTSRB & 0.9546 & 0.9493 & 0.9931 \\
 & MNIST & 0.9916 & 0.9899 & 0.9865 \\
 & CIFAR-10 & 0.8936 & 0.8697 & 0.9252 \\
 & ImageNet & 0.7340 & 0.7130 & 0.8112 \\ \midrule
\multirow{4}{*}{\textbf{Target Model}} & GTSRB & 0.9521 & 0.9485 & 0.9960 \\
 & MNIST & 0.9923 & 0.9799 & 0.9911 \\
 & CIFAR-10 & 0.8959 & 0.8619 & 0.8771 \\
 & ImageNet & 0.8419 & 0.8414 & 0.9620 \\ \bottomrule
\end{tabular}%
}
\end{table}

The attacker and defender each utilize separate datasets to train their respective models, ensuring that the defender has no knowledge of the attacker's training data. Table~\ref{tab:Eset} presents an overview of the average benign accuracy (BA) of benign models, as well as the average BA and average attack success rate (ASR) of backdoored models across datasets. The strong performance exhibited by both benign and backdoored models on these datasets underscores their practical viability for real-world applications. 

\change{\subsubsection{t-SNE Visualization Settings}
We follow the default configuration of the scikit-learn \cite{pedregosa2011scikit} implementation to ensure consistency and reproducibility across experiments. The exact hyperparameter settings are as follows: \texttt{n\_components} = 2, \texttt{perplexity} = 30, \texttt{learning\_rate} = 200, \texttt{n\_iter} = 1000, \texttt{init} = 'pca', \texttt{method} = 'barnes\_hut', \texttt{metric} = 'euclidean', \texttt{early\_exaggeration} = 12.0.
}
\change{We empirically verified that small perturbations in these hyper-parameters (e.g., setting perplexity in the range of 20 to 50 or slightly adjusting the learning rate) do not alter the qualitative distributional patterns observed in our visualizations. }

\subsection{Compare \tool with Learning-based Baselines}
\label{sec:effe_e}

\begin{table*}[tb]
\centering
\caption{Comparison of \tool with baselines (with backdoor detection AUC).}
\label{tab:E1}
\renewcommand{\arraystretch}{0.6}
\resizebox{0.9\linewidth}{!}{%
\begin{tabular}{cccccc|ccc|c}
\toprule
\multirow{3}{*}{\textbf{Dataset}} & \multirow{3}{*}{\textbf{Attack}} & \multirow{3}{*}{\textbf{Approach}} & \multicolumn{3}{c|}{\textbf{Given Model Architectures}} & \multicolumn{3}{c|}{\textbf{Unseen Model Architectures}} & \multirow{3}{*}{All} \\ \cmidrule(lr){4-9}
 &  &  & Simple CNN & SENet18 & ResNet18 & MobileNetV2 & ShuffleNetV2 & EfficientNetB0 &  \\ \midrule
\multirow{8}{*}{GTSRB} & \multirow{3}{*}{BadNets} & MNTD & 0.9678 & 0.8682 & 0.8863 & 0.5678 & 0.5168 & 0.5167 & 0.7455 \\
 &  & ULPs & 0.9236 & 0.6717 & 0.6387 & 0.5514 & 0.5983 & 0.3169 & 0.6532 \\
 &  & \tool & \textbf{0.9940} & \textbf{0.9604} & \textbf{0.9759} & \textbf{0.9438} & \textbf{0.8402} & \textbf{0.6175} & \textbf{0.8698} \\ \cmidrule(l){2-10} 
 & \multirow{3}{*}{Blended} & MNTD & 0.9635 & 0.8755 & 0.8995 & 0.5455 & 0.7667 & 0.4811 & 0.7568 \\
 &  & ULPs & 0.8559 & 0.6824 & 0.6095 & 0.5488 & 0.5380 & 0.3089 & 0.6011 \\
 &  & \tool & \textbf{0.9895} & \textbf{0.9833} & \textbf{0.9887} & \textbf{0.9705} & \textbf{0.8625} & \textbf{0.5704} & \textbf{0.8995} \\ \midrule
\multirow{8}{*}{MNIST} & \multirow{3}{*}{BadNets} & MNTD & \textbf{0.8801} & 0.8040 & 0.8288 & 0.6068 & 0.8185 & 0.5788 & 0.7418 \\
 &  & ULPs & 0.8556 & 0.8120 & \textbf{0.9074} & 0.8128 & 0.7191 & 0.5290 & 0.7110 \\
 &  & \tool & 0.8280 & \textbf{0.8932} & 0.9060 & \textbf{0.8253} & \textbf{0.8223} & \textbf{0.7860} & \textbf{0.8236} \\ \cmidrule(l){2-10} 
 & \multirow{3}{*}{Blended} & MNTD & 0.9316 & \textbf{0.9729} & 0.9367 & 0.6857 & 0.8311 & 0.8216 & 0.8801 \\
 &  & ULPs & 0.6049 & 0.9000 & \textbf{0.9633} & 0.8090 & 0.7024 & 0.4415 & 0.6843 \\
 &  & \tool & \textbf{0.9369} & 0.9684 & 0.9492 & \textbf{0.8961} & \textbf{0.8870} & \textbf{0.8898} & \textbf{0.9206} \\ \midrule
\multirow{8}{*}{CIFAR-10} & \multirow{3}{*}{BadNets} & MNTD & 0.6846 & 0.6890 & 0.7063 & 0.5771 & 0.5838 & 0.5169 & 0.6102 \\
 &  & ULPs & 0.7751 & 0.5549 & 0.5679 & 0.5057 & 0.5133 & 0.5199 & 0.5747 \\
 &  & \tool & \textbf{0.9195} & \textbf{0.9249} & \textbf{0.9430} & \textbf{0.8509} & \textbf{0.8784} & \textbf{0.7452} & \textbf{0.7985} \\ \cmidrule(l){2-10} 
 & \multirow{3}{*}{Blended} & MNTD & 0.7643 & 0.7681 & 0.6795 & 0.4363 & 0.5906 & 0.5602 & 0.6229 \\
 &  & ULPs & 0.8212 & 0.5726 & 0.6108 & 0.4497 & 0.5221 & 0.5206 & 0.5701 \\
 &  & \tool & \textbf{0.9692} & \textbf{0.9647} & \textbf{0.9592} & \textbf{0.8435} & \textbf{0.9561} & \textbf{0.8790} & \textbf{0.9090} \\ \bottomrule
\end{tabular}%
}
\end{table*}

\begin{figure*}[tb]
\centering
\includegraphics[width=\linewidth, scale=0.295]{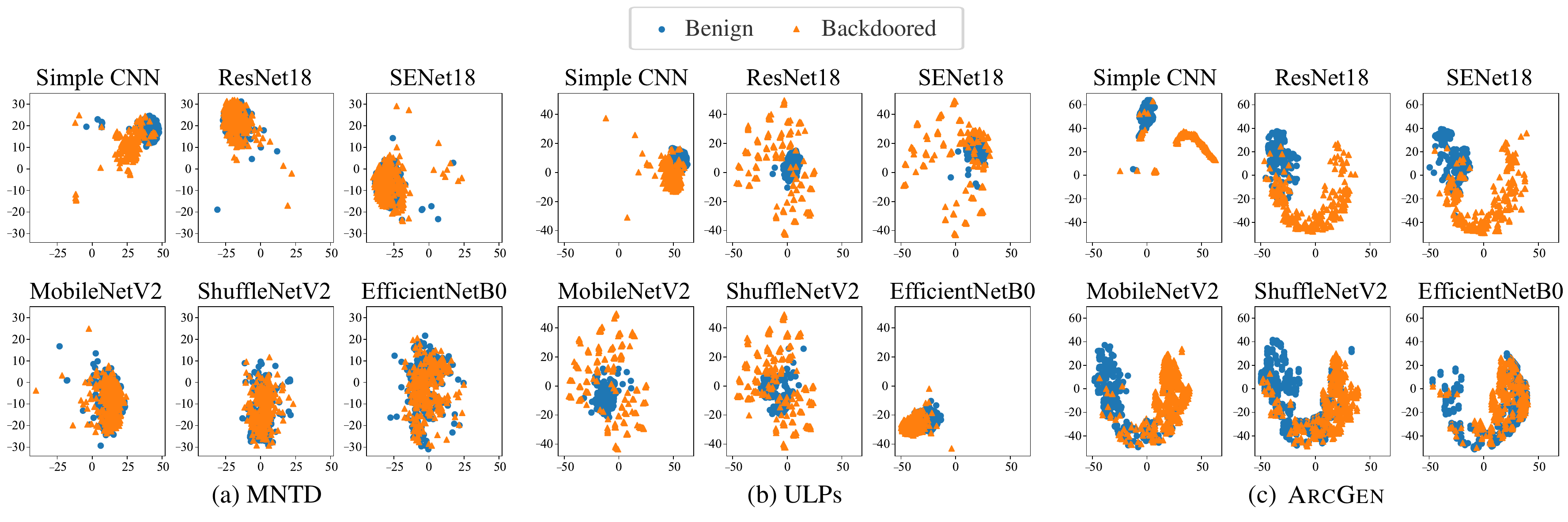}
\caption{The t-SNE visualization of the extracted features by different methods for different model architectures on GTSRB dataset. The color orange represents features for benign models and the blue represents features for backdoored models with BadNets attack. The scatter plot shows the distribution of model features in high-dimensional space.
}
\label{fig:result_feature_dis}
\end{figure*}
Using the experimental setup, we have evaluated the detection performance of \tool{} on different architectures subjected to backdoor attacks. The results are shown in Table~\ref{tab:E1}. The baseline methods show poor performance on unseen model architectures, with significantly lower detection AUCs, sometimes approaching random classification. In contrast, \tool{} demonstrates remarkable performance on unseen model architectures, showcasing substantially higher detection AUCs compared to the baseline methods, with improvements up to 42.5\% under specific conditions. 
On seen model architectures, \tool{}'s detection AUC is either superior or comparable to baseline methods.

To better understand the detection performance differences between \tool{} and baseline methods, we visualize model features extracted by various methods using t-SNE in Figure~\ref{fig:result_feature_dis}. The visualization shows that features extracted by MNTD and ULPs scatter across different locations for various model architectures, leading to poor performance on unseen architectures due to their differing feature distributions from the training phase.
In contrast, features extracted by \tool{} show similar distributions across different model architectures, enabling effective differentiation between benign and backdoored models. Moreover, \tool{} outperforms baseline methods on seen model architectures as well, attributed to its ability to mitigate architectural biases during feature extraction. Mitigating architectural biases allows \tool{} to focus on distinguishing between benign and backdoored models.

We note that while \tool{} generally demonstrates strong generalization to unseen architectures, some performance variation exists across different unseen models. For instance, in Table~\ref{tab:E1}, the detection performance on EfficientNetB0 in the GTSRB dataset is lower than that on other architectures. However, this appears to be an isolated case. On other datasets such as MNIST and CIFAR-10, \tool{} performs consistently well on EfficientNetB0, achieving results comparable to those on seen architectures. Moreover, even in this challenging GTSRB case, \tool{} still outperforms all baselines by a considerable margin (10\%–30\%), underscoring its robust generalization to entirely unseen architectures. We further explore the effect of given architecture combinations on performance in Section~\ref{sec:given}.

\begin{table*}[t]
\centering
\caption{Comparison of \tool and baselines on ImageNet (AUC).}
\label{tab:Eimgnet}
\renewcommand{\arraystretch}{0.6}
\resizebox{0.9\linewidth}{!}{%
\begin{tabular}{cccccccccc|c}
\toprule
\multirow{3}{*}{\textbf{Attack}} & \multirow{3}{*}{\textbf{Approach}} & \multicolumn{3}{c}{\textbf{Given Model Architectures}} & \multicolumn{5}{c|}{\textbf{Unseen Model Architectures}} & \multirow{3}{*}{All} \\ \cmidrule(lr){3-10}
 &  & MobileNetV2 & ResNet18 & \multicolumn{1}{c|}{EfficientNetB0} & SENet18 & ShuffleNetV2 & ViT-B/16 & ViT-B/32 & ViT-L/32 &  \\ \midrule
\multirow{3}{*}{BadNets} & MNTD & 0.8291 & 0.7903 & \multicolumn{1}{c|}{0.8752} & 0.7932 & 0.7654 & 0.7911 & 0.7274 & 0.7334 & 0.7555 \\
 & ULPs & 0.4800 & 0.4874 & \multicolumn{1}{c|}{0.6403} & 0.4932 & 0.3281 & 0.4992 & 0.4912 & 0.6572 & 0.5122 \\
 & \tool & \textbf{0.9021} & \textbf{0.8787} & \multicolumn{1}{c|}{\textbf{0.9438}} & \textbf{0.8842} & \textbf{0.9392} & \textbf{0.8103} & \textbf{0.7956} & \textbf{0.8807} & \textbf{0.8360} \\ \midrule
\multirow{3}{*}{Blended} & MNTD & 0.7933 & 0.5809 & \multicolumn{1}{c|}{0.8742} & 0.5951 & 0.6763 & 0.8593 & 0.9235 & 0.8545 & 0.6979 \\
 & ULPs & 0.6137 & 0.5266 & \multicolumn{1}{c|}{0.6291} & 0.5144 & 0.6018 & 0.5156 & 0.5006 & 0.6721 & 0.5593 \\
 & \tool & \textbf{0.8786} & \textbf{0.7981} & \multicolumn{1}{c|}{\textbf{0.9563}} & \textbf{0.7564} & \textbf{0.8451} & \textbf{0.8906} & \textbf{0.9568} & \textbf{0.9452} & \textbf{0.8029} \\ \bottomrule
\end{tabular}%
}
\end{table*}
\textit{Effectiveness on ImageNet.}
We evaluate \tool's effectiveness on challenging tasks using the ImageNet dataset \cite{ILSVRC15} and advanced transformer-based models. 
Since Simple CNN performs poorly on this task, we remove it from the given architecture set. Instead, we randomly select three architectures from the pool: ResNet18, MobileNetV2, and EfficientNetB0. For unseen models, we use SENet18, ShuffleNetV2, ViT-Base/16, ViT-Base/32, and ViT-Large/32. In this evaluation, the attacker trains 64 models for each attack and architecture, employing both BadNets and Blended attacks.

The results are presented in Table~\ref{tab:Eimgnet}. Like other datasets, \tool shows excellent generalization performance on unseen model architectures and outperforms baseline methods on both unseen and seen model architectures. Additionally, \tool demonstrates outstanding detection performance on transformer-based models, notably surpassing the baselines.

\subsection{Compare \tool with MM-BD}
\begin{table*}[tb]
\centering
\caption{Comparison of \tool with MM-BD (with backdoor detection AUC).}
\label{tab:EMMBD}
\begin{minipage}{\linewidth}
\centering
\renewcommand{\arraystretch}{0.6}
\resizebox{0.8\linewidth}{!}{%
\begin{tabular}{@{}cccccccc@{}}
\toprule
\textbf{Dataset} & \textbf{Approach} & \textbf{Simple CNN} & \textbf{SENet18} & \textbf{ResNet18} & \textbf{MobileNetV2} & \textbf{ShuffleNetV2} & \textbf{EfficientNetB0} \\ \midrule
\multirow{2}{*}{{GTSRB}} & MM-BD & 0.6172 & 0.9492 & 0.9570 & 0.7891 & 0.7422 & 0.4941 \\ 
& \tool & \textbf{0.9940} & \textbf{0.9604} & \textbf{0.9759} & \textbf{0.9438} & \textbf{0.8402} & \textbf{0.6175} \\ \midrule
\multirow{2}{*}{{MNIST}} & MM-BD & 0.5957 & 0.8281 & 0.7422 & 0.8086 & 0.5996 & 0.7363 \\ 
& \tool & \textbf{0.8280} & \textbf{0.8932} & \textbf{0.9060} & \textbf{0.8253} & \textbf{0.8223} & \textbf{0.7860} \\ \midrule
\multirow{2}{*}{{CIFAR-10}} & MM-BD & 0.5137 & \textbf{0.9355} & 0.9395 & 0.5820 & 0.7949 & 0.4492 \\
& \tool & \textbf{0.9195} & 0.9249 & \textbf{0.9430} & \textbf{0.8509} & \textbf{0.8784} & \textbf{0.7452} \\ \bottomrule
\end{tabular}
}
\end{minipage}

\vspace{0.1cm} 

\begin{minipage}{\linewidth}
\centering
\renewcommand{\arraystretch}{0.8}
\resizebox{0.8\linewidth}{!}{%
\begin{tabular}{@{}cccccccccc@{}}
\toprule
\textbf{Dataset} & \textbf{Approach} & \textbf{MobileNetV2} & \textbf{ResNet18} & \textbf{EfficientNetB0} & \textbf{SENet18} & \textbf{ShuffleNetV2} & \textbf{ViT-B/16} & \textbf{ViT-B/32} & \textbf{ViT-L/32} \\ \midrule
\multirow{2}{*}{{ImageNet}} & MM-BD & 0.8750 & 0.7813 & 0.7500 & \textbf{0.9375} & 0.7813 & 0.4812 & 0.5313 & 0.4688 \\
& \tool & \textbf{0.9021} & \textbf{0.8787} & \textbf{0.9438} & 0.8842 & \textbf{0.9392} & \textbf{0.8103} & \textbf{0.7956} & \textbf{0.8807} \\ \bottomrule
\end{tabular}
}
\end{minipage}
\end{table*}

Maximum-Margin-based Backdoor Detection (MM-BD) \cite{wang2023mm} is a black-box model-based detection method that is recently introduced. MM-BD does not involve a feature learning process, so it does not distinguish between seen and unseen architectures. Therefore, we compare its performance with \tool in a separate table. This method calculates a maximum margin detection threshold for each class, defined as the maximum difference between the logit value of a target class and the maximum logit value among all other classes in the classifier's logit space. An abnormally large maximum margin value indicates the presence of a backdoor.

Following the defender settings in Section~\ref{sec:effe_e}, we evaluate the detection effectiveness of \tool and MM-BD against the BadNets attack. As shown in Table~\ref{tab:EMMBD}, \tool outperforms MM-BD across various architectures and datasets. 
The average performance improvement on GTSRB, MNIST, CIFAR10, and ImageNet are 13.11\%, 12.51\%, 17.45\%, and 17.85\%, respectively.
We attribute this significant performance improvement to \tool's ability to automatically learn the detection function, avoiding human biases in design.

\begin{table*}[t]
\centering
\caption{\change{Detection AUC when relaxing the training distribution assumption. Detection approaches are trained on models using one set of 10 ImageNet classes and evaluated on models trained on a disjoint set of 10 different classes.}}
\label{tab:cross_dist}
\resizebox{0.9\textwidth}{!}{%
\begin{tabular}{ccccccccccc}
\toprule
\multirow{2}{*}{\textbf{Attack}} & \multirow{2}{*}{\textbf{Approach}} & \multicolumn{3}{c}{\textbf{Given Model Architectures}} & \multicolumn{5}{c}{\textbf{Unseen Model Architectures}} & \multirow{2}{*}{All} \\ \cmidrule(lr){3-10}
 &  & MobileNetV2 & ResNet18 & EfficientNetB0 & SENet18 & ShuffleNetV2 & ViT-B/16 & ViT-B/32 & ViT-L/32 &  \\ \midrule
\multirow{2}{*}{BadNets} & MNTD & 0.8562 & 0.7253 & 0.8713 & 0.7595 & 0.5692 & 0.6764 & 0.6713 & 0.6366 & 0.6261 \\
 & \tool{} & 0.9002 & 0.9038 & 0.8755 & 0.8021 & 0.8256 & 0.7253 & 0.7336 & 0.8083 & 0.7605 \\
\multirow{2}{*}{Blended} & MNTD & 0.8085 & 0.6315 & 0.7739 & 0.5883 & 0.6764 & 0.7365 & 0.6913 & 0.6366 & 0.6860 \\
 & \tool{} & 0.8997 & 0.7964 & 0.9586 & 0.7849 & 0.8645 & 0.7756 & 0.7143 & 0.6798 & 0.7356 \\ \bottomrule 
\end{tabular}%
}
\end{table*}

\change{\subsection{Evaluation under Training Distribution Shift}}
\label{sec:exp-cross-dist}
\change{While our method treats the target model as a black box, we assume the defender has access to a surrogate dataset drawn from a similar distribution, consistent with prior black-box defenses such as MNTD \cite{Xu2021a} and ULPs \cite{Kolouri2020}. To evaluate the robustness of our method to training distribution mismatch, we design an experiment simulating a practical scenario where the training distribution of the target model is unknown. Specifically, we train the \tool{} using ten disjoint classes from ImageNet that do not overlap with those used to train the target models. Table~\ref{tab:cross_dist} presents the AUC results across various architectures and two representative attack types (BadNets and Blended).}

\change{As shown in Table~\ref{tab:cross_dist}, \tool{} consistently outperforms a strong baseline (MNTD) across both seen and unseen architectures, even under a distribution shift. Specifically, \tool{} achieves up to 14\% higher AUC in detecting backdoors, indicating its robustness and generalization ability in practical deployment scenarios. These results demonstrate that \tool{} remains effective even when precise knowledge of the training distribution is unavailable, reinforcing its practicality for real-world black-box settings.}

\subsection{Performance on Unseen Stealthy Attack}

To further evaluate the generalization performance of our method, we employ three stealthy attacks that are unseen in the detector training phase, and they are SIG \cite{Barni2019}, Bppattack \cite{wang2022bppattack}, and WaNet \cite{nguyen2021wanet}. These attacks use triggers with special shapes following certain patterns. In addition to the special triggers, WaNet proposes a novel training mode called ``noise" mode, which makes such models undetectable by machine defenders \cite{nguyen2021wanet}. Bppattack leverages contrastive supervised learning and adversarial training to make the attack hard to detect \cite{wang2022bppattack}. 
Specifically, we generate 256 target backdoored models for each architecture using SIG attacks on the MNIST dataset. Additionally, we have trained 64 models for each attack and architecture using Bppattack and WaNet attacks on the GTSRB dataset. The target benign models described in Section~\ref{sec:effe_e} are utilized for these experiments, and each experiment is repeated five times for evaluation.

\begin{table*}[tb]
\centering
\caption{Comparison of different detection methods on unseen stealthy backdoor attack (AUC).}
\label{tab:Eotherattack}
\renewcommand{\arraystretch}{0.6}
\resizebox{0.8\linewidth}{!}{%
\begin{tabular}{cccc|ccc}
\toprule
\multirow{3}{*}{\textbf{Attack}} & \multirow{3}{*}{\textbf{Approach}} & \multicolumn{2}{c|}{\textbf{Given Model Architectures}} & \multicolumn{3}{c}{\textbf{Unseen Model Architectures}} \\ \cmidrule(l){3-7} 
 &  & SENet18 & ResNet18 & MobileNetV2 & ShuffleNetV2 & EfficientNetB0 \\ \midrule
\multirow{3}{*}{SIG} & MNTD & \textbf{0.9763} & 0.9325 & 0.6926 & 0.8165 & 0.8197 \\
 & ULPs & 0.8307 & 0.7654 & 0.5606 & 0.7647 & 0.4151 \\
 & \tool & 0.9703 & \textbf{0.9503} & \textbf{0.8973} & \textbf{0.8801} & \textbf{0.9023} \\ \midrule
\multirow{3}{*}{WaNet} & MNTD & \textbf{0.9990} & \textbf{1.0000} & 0.8325 & 0.6285 & 0.6119 \\
 & ULPs & 0.9265 & 0.8780 & 0.5349 & 0.5728 & 0.3105 \\
 & \tool & 0.9657 & 0.9942 & \textbf{0.9206} & \textbf{0.8015} & \textbf{0.7205} \\ \midrule
\multirow{3}{*}{BppAttack} & MNTD & 0.9860 & 0.9800 & 0.8771 & 0.7137 & 0.5885 \\
 & ULPs & 0.9232 & 0.8436 & 0.5032 & 0.5777 & 0.3423 \\
 & \tool & \textbf{0.9967} & \textbf{0.9963} & \textbf{0.9352} & \textbf{0.7664} & \textbf{0.7360} \\ \bottomrule
\end{tabular}
}
\end{table*}
Table~\ref{tab:Eotherattack} presents the comparison of various defense methods on these stealthy attack methods. \tool outperforms MNTD and ULPs significantly across all unseen model architectures and exhibits comparable or superior performance on given model architectures, demonstrating its effectiveness in identifying backdoor threats, even if we are completely unaware of such attacks during learning. 

\subsection{Influence of Different Given Architectures}
\label{sec:given}
\begin{table*}[t]
\centering
\caption{The influence of different combinations of seen architectures. This table shows the detection AUC of each approach on different model architectures on the GTSRB dataset. We evaluate eight different architecture combinations. {\colorbox[HTML]{9AFF99}{Green}} and {\colorbox[HTML]{FFCCC9}{red}} represent the results on the seen and unseen architectures, respectively.}
\label{tab:Ediffseen}
\renewcommand{\arraystretch}{0.6}
\begin{tabular}{cccccccc}
\toprule
\multirow{4}{*}{1} & \textbf{Approach} & \textbf{\cellcolor[HTML]{9AFF99}Simple CNN} & \textbf{\cellcolor[HTML]{9AFF99}SENet18} & \textbf{\cellcolor[HTML]{FFCCC9}ResNet18} & \textbf{\cellcolor[HTML]{FFCCC9}MobileNetV2} & \textbf{\cellcolor[HTML]{FFCCC9}ShuffeNetV2} & \textbf{\cellcolor[HTML]{FFCCC9}EfficientNetB0} \\ \cmidrule(l){2-8} 
 & \textbf{MNTD} & {\cellcolor[HTML]{9AFF99}0.9830} & {\cellcolor[HTML]{9AFF99}0.8426} & {\cellcolor[HTML]{FFCCC9}0.6933} & {\cellcolor[HTML]{FFCCC9}0.5464} & {\cellcolor[HTML]{FFCCC9}0.4708} & {\cellcolor[HTML]{FFCCC9}0.4651} \\
 & \textbf{ULPs} & {\cellcolor[HTML]{9AFF99}0.9280} & {\cellcolor[HTML]{9AFF99}0.6927} & {\cellcolor[HTML]{FFCCC9}0.5853} & {\cellcolor[HTML]{FFCCC9}0.5297} & {\cellcolor[HTML]{FFCCC9}0.5541} & {\cellcolor[HTML]{FFCCC9}0.2878} \\
 & \textbf{\tool} & \textbf{\cellcolor[HTML]{9AFF99}0.9933} & \textbf{\cellcolor[HTML]{9AFF99}0.9508} & \textbf{\cellcolor[HTML]{FFCCC9}0.8063} & \textbf{\cellcolor[HTML]{FFCCC9}0.7622} & \textbf{\cellcolor[HTML]{FFCCC9}0.6522} & \textbf{\cellcolor[HTML]{FFCCC9}0.4878} \\ \midrule
\multirow{4}{*}{2} & \textbf{Approach} & \textbf{\cellcolor[HTML]{9AFF99}Simple CNN} & \textbf{\cellcolor[HTML]{9AFF99}MobileNetV2} & \textbf{\cellcolor[HTML]{FFCCC9}SENet18} & \textbf{\cellcolor[HTML]{FFCCC9}ResNet18} & \textbf{\cellcolor[HTML]{FFCCC9}ShuffeNetV2} & \textbf{\cellcolor[HTML]{FFCCC9}EfficientNetB0} \\ \cmidrule(l){2-8} 
 & \textbf{MNTD} & \textbf{\cellcolor[HTML]{9AFF99}0.9801} & {\cellcolor[HTML]{9AFF99}0.7965} & {\cellcolor[HTML]{FFCCC9}0.6456} & {\cellcolor[HTML]{FFCCC9}0.7321} & {\cellcolor[HTML]{FFCCC9}0.6802} & {\cellcolor[HTML]{FFCCC9}0.5422} \\
 & \textbf{ULPs} & {\cellcolor[HTML]{9AFF99}0.9263} & {\cellcolor[HTML]{9AFF99}0.5991} & {\cellcolor[HTML]{FFCCC9}0.4660} & {\cellcolor[HTML]{FFCCC9}0.6088} & {\cellcolor[HTML]{FFCCC9}0.5919} & {\cellcolor[HTML]{FFCCC9}0.5903} \\
 & \textbf{\tool} & {\cellcolor[HTML]{9AFF99}0.9175} & \textbf{\cellcolor[HTML]{9AFF99}0.9869} & \textbf{\cellcolor[HTML]{FFCCC9}0.9702} & \textbf{\cellcolor[HTML]{FFCCC9}0.9709} & \textbf{\cellcolor[HTML]{FFCCC9}0.9403} & \textbf{\cellcolor[HTML]{FFCCC9}0.7934} \\ \midrule
\multirow{4}{*}{3} & \textbf{Approach} & \textbf{\cellcolor[HTML]{9AFF99}ResNet18} & \textbf{\cellcolor[HTML]{9AFF99}MobileNetV2} & \textbf{\cellcolor[HTML]{FFCCC9}Simple CNN} & \textbf{\cellcolor[HTML]{FFCCC9}SENet18} & \textbf{\cellcolor[HTML]{FFCCC9}ShuffeNetV2} & \textbf{\cellcolor[HTML]{FFCCC9}EfficientNetB0} \\ \cmidrule(l){2-8} 
 & \textbf{MNTD} & {\cellcolor[HTML]{9AFF99}0.9157} & {\cellcolor[HTML]{9AFF99}0.8016} & {\cellcolor[HTML]{FFCCC9}0.5464} & {\cellcolor[HTML]{FFCCC9}0.7223} & {\cellcolor[HTML]{FFCCC9}0.7219} & {\cellcolor[HTML]{FFCCC9}0.6340} \\
 & \textbf{ULPs} & {\cellcolor[HTML]{9AFF99}0.7794} & {\cellcolor[HTML]{9AFF99}0.7908} & {\cellcolor[HTML]{FFCCC9}0.4743} & {\cellcolor[HTML]{FFCCC9}0.6510} & {\cellcolor[HTML]{FFCCC9}0.8177} & {\cellcolor[HTML]{FFCCC9}0.7586} \\
 & \textbf{\tool} & \textbf{\cellcolor[HTML]{9AFF99}0.9839} & \textbf{\cellcolor[HTML]{9AFF99}0.9761} & \textbf{\cellcolor[HTML]{FFCCC9}0.8321} & \textbf{\cellcolor[HTML]{FFCCC9}0.8225} & \textbf{\cellcolor[HTML]{FFCCC9}0.9360} & \textbf{\cellcolor[HTML]{FFCCC9}0.8322} \\ \midrule
\multirow{4}{*}{4} & \textbf{Approach} & \textbf{\cellcolor[HTML]{9AFF99}Simple CNN} & \textbf{\cellcolor[HTML]{9AFF99}SENet18} & \textbf{\cellcolor[HTML]{9AFF99}ResNet18} & \textbf{\cellcolor[HTML]{FFCCC9}MobileNetV2} & \textbf{\cellcolor[HTML]{FFCCC9}ShuffeNetV2} & \textbf{\cellcolor[HTML]{FFCCC9}EfficientNetB0} \\ \cmidrule(l){2-8} 
 & \textbf{MNTD} & {\cellcolor[HTML]{9AFF99}0.9678} & {\cellcolor[HTML]{9AFF99}0.8682} & {\cellcolor[HTML]{9AFF99}0.8863} & {\cellcolor[HTML]{FFCCC9}0.5678} & {\cellcolor[HTML]{FFCCC9}0.5168} & {\cellcolor[HTML]{FFCCC9}0.5167} \\
 & \textbf{ULPs} & {\cellcolor[HTML]{9AFF99}0.9236} & {\cellcolor[HTML]{9AFF99}0.6717} & {\cellcolor[HTML]{9AFF99}0.6387} & {\cellcolor[HTML]{FFCCC9}0.5514} & {\cellcolor[HTML]{FFCCC9}0.5983} & {\cellcolor[HTML]{FFCCC9}0.3169} \\
 & \textbf{\tool} & \textbf{\cellcolor[HTML]{9AFF99}0.9940} & \textbf{\cellcolor[HTML]{9AFF99}0.9604} & \textbf{\cellcolor[HTML]{9AFF99}0.9759} & \textbf{\cellcolor[HTML]{FFCCC9}0.9438} & \textbf{\cellcolor[HTML]{FFCCC9}0.8402} & \textbf{\cellcolor[HTML]{FFCCC9}0.6175} \\ \midrule
\multirow{4}{*}{5} & \textbf{Approach} & \textbf{\cellcolor[HTML]{9AFF99}Simple CNN} & \textbf{\cellcolor[HTML]{9AFF99}MobileNetV2} & \textbf{\cellcolor[HTML]{9AFF99}ResNet18} & \textbf{\cellcolor[HTML]{FFCCC9}SENet18} & \textbf{\cellcolor[HTML]{FFCCC9}ShuffeNetV2} & \textbf{\cellcolor[HTML]{FFCCC9}EfficientNetB0} \\ \cmidrule(l){2-8} 
 & \textbf{MNTD} & {\cellcolor[HTML]{9AFF99}0.9766} & {\cellcolor[HTML]{9AFF99}0.8620} & {\cellcolor[HTML]{9AFF99}0.9305} & {\cellcolor[HTML]{FFCCC9}0.7200} & {\cellcolor[HTML]{FFCCC9}0.7438} & {\cellcolor[HTML]{FFCCC9}0.6016} \\
 & \textbf{ULPs} & {\cellcolor[HTML]{9AFF99}0.9319} & {\cellcolor[HTML]{9AFF99}0.6330} & {\cellcolor[HTML]{9AFF99}0.6803} & {\cellcolor[HTML]{FFCCC9}0.5163} & {\cellcolor[HTML]{FFCCC9}0.6303} & {\cellcolor[HTML]{FFCCC9}0.5198} \\
 & \textbf{\tool} & \textbf{\cellcolor[HTML]{9AFF99}0.9885} & \textbf{\cellcolor[HTML]{9AFF99}0.9779} & \textbf{\cellcolor[HTML]{9AFF99}0.9739} & \textbf{\cellcolor[HTML]{FFCCC9}0.8653} & \textbf{\cellcolor[HTML]{FFCCC9}0.9013} & \textbf{\cellcolor[HTML]{FFCCC9}0.7572} \\ \midrule
\multirow{4}{*}{6} & \textbf{Approach} & \textbf{\cellcolor[HTML]{9AFF99}Simple CNN} & \textbf{\cellcolor[HTML]{9AFF99}SENet18} & \textbf{\cellcolor[HTML]{9AFF99}MobileNetV2} & \textbf{\cellcolor[HTML]{FFCCC9}ResNet18} & \textbf{\cellcolor[HTML]{FFCCC9}ShuffeNetV2} & \textbf{\cellcolor[HTML]{FFCCC9}EfficientNetB0} \\ \cmidrule(l){2-8} 
 & \textbf{MNTD} & {\cellcolor[HTML]{9AFF99}0.9803} & {\cellcolor[HTML]{9AFF99}0.8929} & {\cellcolor[HTML]{9AFF99}0.8624} & {\cellcolor[HTML]{FFCCC9}0.8464} & {\cellcolor[HTML]{FFCCC9}0.7409} & {\cellcolor[HTML]{FFCCC9}0.5772} \\
 & \textbf{ULPs} & {\cellcolor[HTML]{9AFF99}0.9454} & {\cellcolor[HTML]{9AFF99}0.7033} & {\cellcolor[HTML]{9AFF99}0.6119} & {\cellcolor[HTML]{FFCCC9}0.6421} & {\cellcolor[HTML]{FFCCC9}0.6245} & {\cellcolor[HTML]{FFCCC9}0.4081} \\
 & \textbf{\tool} & \textbf{\cellcolor[HTML]{9AFF99}0.9978} & \textbf{\cellcolor[HTML]{9AFF99}0.9711} & \textbf{\cellcolor[HTML]{9AFF99}0.9848} & \textbf{\cellcolor[HTML]{FFCCC9}0.9617} & \textbf{\cellcolor[HTML]{FFCCC9}0.9129} & \textbf{\cellcolor[HTML]{FFCCC9}0.7328} \\ \midrule
\multirow{4}{*}{7} & \textbf{Approach} & \textbf{\cellcolor[HTML]{9AFF99}MobileNetV2} & \textbf{\cellcolor[HTML]{9AFF99}SENet18} & \textbf{\cellcolor[HTML]{9AFF99}ResNet18} & \textbf{\cellcolor[HTML]{FFCCC9}Simple CNN} & \textbf{\cellcolor[HTML]{FFCCC9}ShuffeNetV2} & \textbf{\cellcolor[HTML]{FFCCC9}EfficientNetB0} \\ \cmidrule(l){2-8} 
 & \textbf{MNTD} & {\cellcolor[HTML]{9AFF99}0.8815} & {\cellcolor[HTML]{9AFF99}0.9111} & {\cellcolor[HTML]{9AFF99}0.9516} & {\cellcolor[HTML]{FFCCC9}0.5319} & {\cellcolor[HTML]{FFCCC9}0.8159} & {\cellcolor[HTML]{FFCCC9}0.6802} \\
 & \textbf{ULPs} & {\cellcolor[HTML]{9AFF99}0.7786} & {\cellcolor[HTML]{9AFF99}0.8179} & {\cellcolor[HTML]{9AFF99}0.7917} & {\cellcolor[HTML]{FFCCC9}0.5786} & {\cellcolor[HTML]{FFCCC9}0.8079} & {\cellcolor[HTML]{FFCCC9}0.5301} \\
 & \textbf{\tool} & \textbf{\cellcolor[HTML]{9AFF99}0.9877} & \textbf{\cellcolor[HTML]{9AFF99}0.9746} & \textbf{\cellcolor[HTML]{9AFF99}0.9867} & \textbf{\cellcolor[HTML]{FFCCC9}0.8949} & \textbf{\cellcolor[HTML]{FFCCC9}0.9563} & \textbf{\cellcolor[HTML]{FFCCC9}0.9018} \\ \midrule
\multirow{4}{*}{8} & \textbf{Approach} & \textbf{\cellcolor[HTML]{9AFF99}Simple CNN} & \textbf{\cellcolor[HTML]{9AFF99}SENet18} & \textbf{\cellcolor[HTML]{9AFF99}ResNet18} & \textbf{\cellcolor[HTML]{9AFF99}MobileNetV2} & \textbf{\cellcolor[HTML]{FFCCC9}ShuffeNetV2} & \textbf{\cellcolor[HTML]{FFCCC9}EfficientNetB0} \\ \cmidrule(l){2-8} 
 & \textbf{MNTD} & {\cellcolor[HTML]{9AFF99}0.9791} & {\cellcolor[HTML]{9AFF99}0.8794} & {\cellcolor[HTML]{9AFF99}0.9425} & {\cellcolor[HTML]{9AFF99}0.8705} & {\cellcolor[HTML]{FFCCC9}0.7484} & {\cellcolor[HTML]{FFCCC9}0.5635} \\
 & \textbf{ULPs} & {\cellcolor[HTML]{9AFF99}0.9495} & {\cellcolor[HTML]{9AFF99}0.7250} & {\cellcolor[HTML]{9AFF99}0.7155} & {\cellcolor[HTML]{9AFF99}0.6321} & {\cellcolor[HTML]{FFCCC9}0.6795} & {\cellcolor[HTML]{FFCCC9}0.4394} \\
 & \textbf{\tool} & \textbf{\cellcolor[HTML]{9AFF99}0.9943} & \textbf{\cellcolor[HTML]{9AFF99}0.9671} & \textbf{\cellcolor[HTML]{9AFF99}0.9803} & \textbf{\cellcolor[HTML]{9AFF99}0.9857} & \textbf{\cellcolor[HTML]{FFCCC9}0.9182} & \textbf{\cellcolor[HTML]{FFCCC9}0.7303} \\ \bottomrule
\end{tabular}
\end{table*}

In this section, we investigate the performance of \tool across various combinations of given model architectures. 
We evaluate our method using the AUC metric, calculated by averaging results from five runs. 
The results in Table \ref{tab:Ediffseen} demonstrate that \tool consistently outperforms the baseline across various combinations of given model architectures, demonstrating the robustness of our method.

Further, generally, incorporating more architectures tends to enhance detection performance. As demonstrated in Table \ref{tab:Ediffseen}, the average performance is better when using three architectures than two, based on the results of combination No.123 versus No.4567. However, adding more architectures does not always yield better results, and the specific combinations of architectures also play a significant role. For example, combination No.4, which had more seen architectures than combination No.3, showed a notable decrease in detection performance for ShuffleNetV2 and EfficientNetB0. This phenomenon will be further explored in future work.

\subsection{Ablation Studies}
\label{sec:ablation_exp}
\begin{table*}[tb]
\centering
\caption{\change{The detection AUC of \tool and its variants with specific components removed. The results demonstrate the effectiveness of the distribution-level alignment loss ($\mathcal{L}_{dla}$), sample-level alignment loss ($\mathcal{L}_{sla}$), architecture randomization (Arch Rand), and alignment layers (ALs).}}

\label{tab:E2}
\renewcommand{\arraystretch}{0.6}
\begin{tabular}{ccccc|ccc|c}
\toprule
\multirow{2}{*}{\textbf{Attack}} & \multirow{2}{*}{\textbf{Approach}} & \multicolumn{3}{c|}{\textbf{Given Model Architectures}} & \multicolumn{3}{c|}{\textbf{Unseen Model Architectures}} & \multirow{2}{*}{All} \\ \cmidrule(lr){3-8}
 &  & Simple CNN & SENet18 & ResNet18 & MobileNetV2 & ShuffleNetV2 & EfficientNetB0 &  \\ \midrule
\multirow{6}{*}{BadNets} 
 & \tool & 0.9940 & 0.9604 & 0.9759 & \textbf{0.9438} & \textbf{0.8402} & 0.6175 & \textbf{0.8698} \\
 & w/o $\mathcal{L}_{sla}$ & \textbf{0.9954} & 0.9601 & \textbf{0.9784} & 0.9349 & 0.8170 & \textbf{0.6337} & 0.8600 \\
 & w/o $\mathcal{L}_{dla}$ & 0.9950 & 0.9578 & 0.9689 & 0.8915 & 0.7710 & 0.6253 & 0.8380 \\
 & \change{w/o $\mathcal{L}_{sla}$ \& $\mathcal{L}_{dla}$} & \change{0.9687} & \change{0.8686} & \change{0.8540} & \change{0.5913} & \change{0.5051} & \change{0.4945} & \change{0.7356} \\
 & w/o Arch Rand & 0.9951 & \textbf{0.9612} & 0.9664 & 0.9194 & 0.7895 & 0.6190 & 0.8415 \\
 & w/o ALs & 0.5078 & 0.5176 & 0.5293 & 0.5918 & 0.5587 & 0.5236 & 0.5381 \\  \midrule
\multirow{6}{*}{Blended} 
 & \tool & 0.9895 & 0.9833 & 0.9887 & 0.9705 & \textbf{0.8625} & 0.5704 & \textbf{0.8995} \\
 & w/o $\mathcal{L}_{sla}$ & 0.9890 & 0.9715 & \textbf{0.9889} & \textbf{0.9780} & 0.8115 & \textbf{0.5889} & 0.8867 \\
 & w/o $\mathcal{L}_{dla}$ & 0.9909 & \textbf{0.9876} & 0.9958 & 0.9275 & 0.8062 & 0.5888 & 0.8555 \\
 & \change{w/o $\mathcal{L}_{sla}$ \& $\mathcal{L}_{dla}$} & \change{0.9664} & \change{0.8798} & \change{0.8620} & \change{0.5906} & \change{0.5047} & \change{0.4929} & \change{0.7360} \\
 & w/o Arch Rand & \textbf{0.9940} & 0.9848 & 0.9882 & 0.9723 & 0.8179 & 0.5741 & 0.8827 \\
 & w/o ALs & 0.6245 & 0.5273 & 0.5703 & 0.6602 & 0.5236 & 0.5097 & 0.5692 \\ 
 \bottomrule
\end{tabular}%
\end{table*}
We conduct ablation studies on GTSRB to analyze the impact of different components on the detection performance of \tool. 
Specifically, we evaluate the effect of sample-level alignment loss, architecture randomization, or alignment layers on the detection performance by removing them. When removing alignment layers, fairness is ensured by relocating the alignment layers' structure from the feature extractor to the detector. 
Each experiment is repeated five times, and the average results are reported in Table~\ref{tab:E2}.

We observe that both alignment losses contribute to the improved performance of our method. The sample-level and distribution-level alignment losses enhance the overall performance by approximately 1.2\% and 4.5\%, respectively.
Moreover, the randomization of architecture also improves overall performance.

\begin{figure}[tb]
\centering
\includegraphics[width=0.8\linewidth]{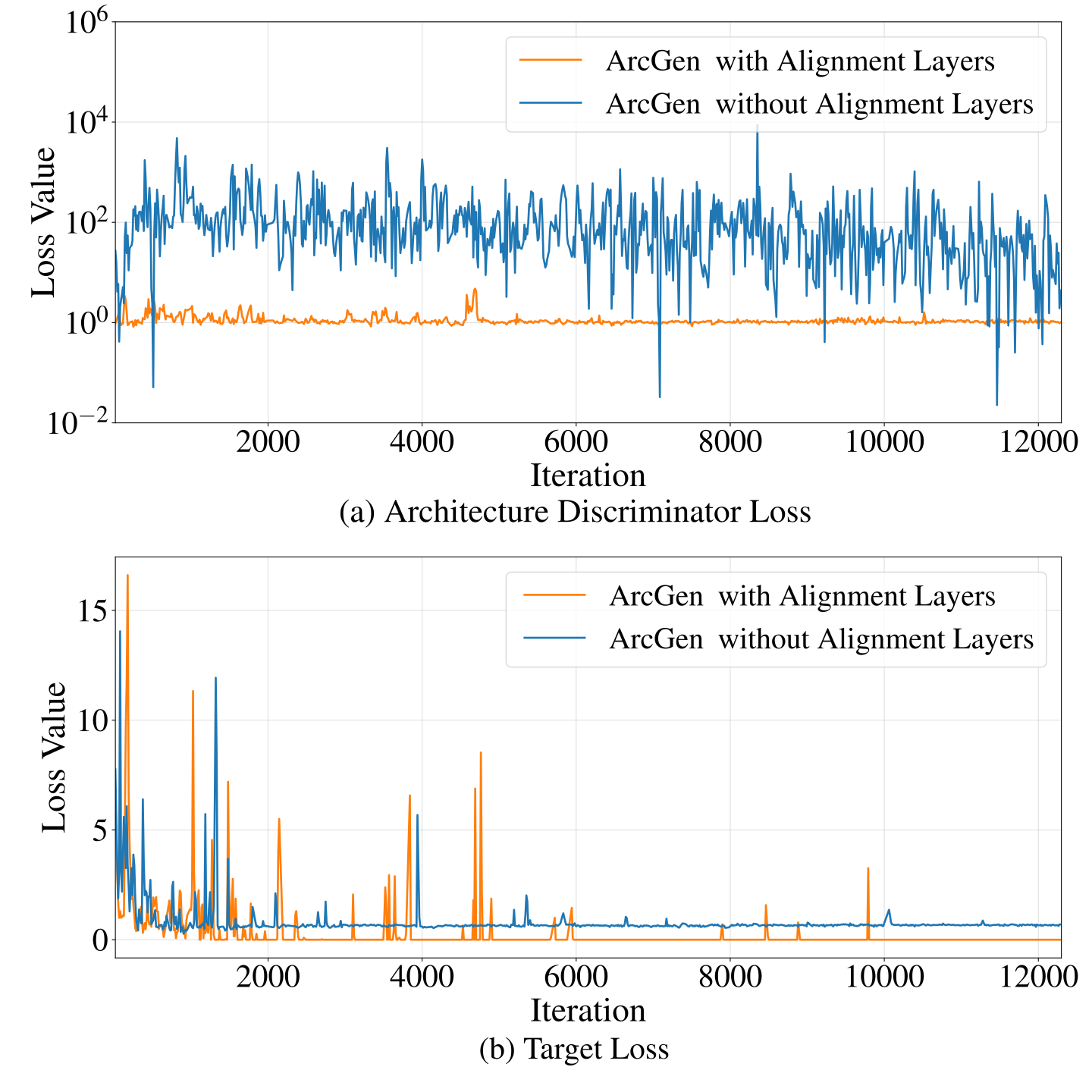}
\caption{
The importance of alignment layers. When we remove these layers, the discriminator loss struggles to converge, and the target loss cannot be effectively optimized.
}
\label{fig:result_withoutal}
\end{figure}

We also observe that removing alignment layers results in a significant drop in the detection performance. 
To analyze the root cause, we compare the adversarial learning process with and without alignment layers. 
Figure~\ref{fig:result_withoutal} shows that during the adversarial learning process, the target loss $\mathcal{L}_t$ does not converge to smaller values, and the architecture discriminator loss $\mathcal{L}_{ad}$ struggles to converge effectively when alignment layers are removed. 
This observation supports our earlier discussion in Section~\ref{sec:allayer}: When model outputs are used as model features directly, such features are strongly associated with the model architecture. This hampers the extraction of architecture-invariant features. Therefore, the alignment layers are crucial for our method to achieve high detection performance.

\subsection{The Effect of Hyper-parameters}
\label{sec:hyper}
\begin{figure}[t]
\centering
\includegraphics[width=0.8\linewidth]{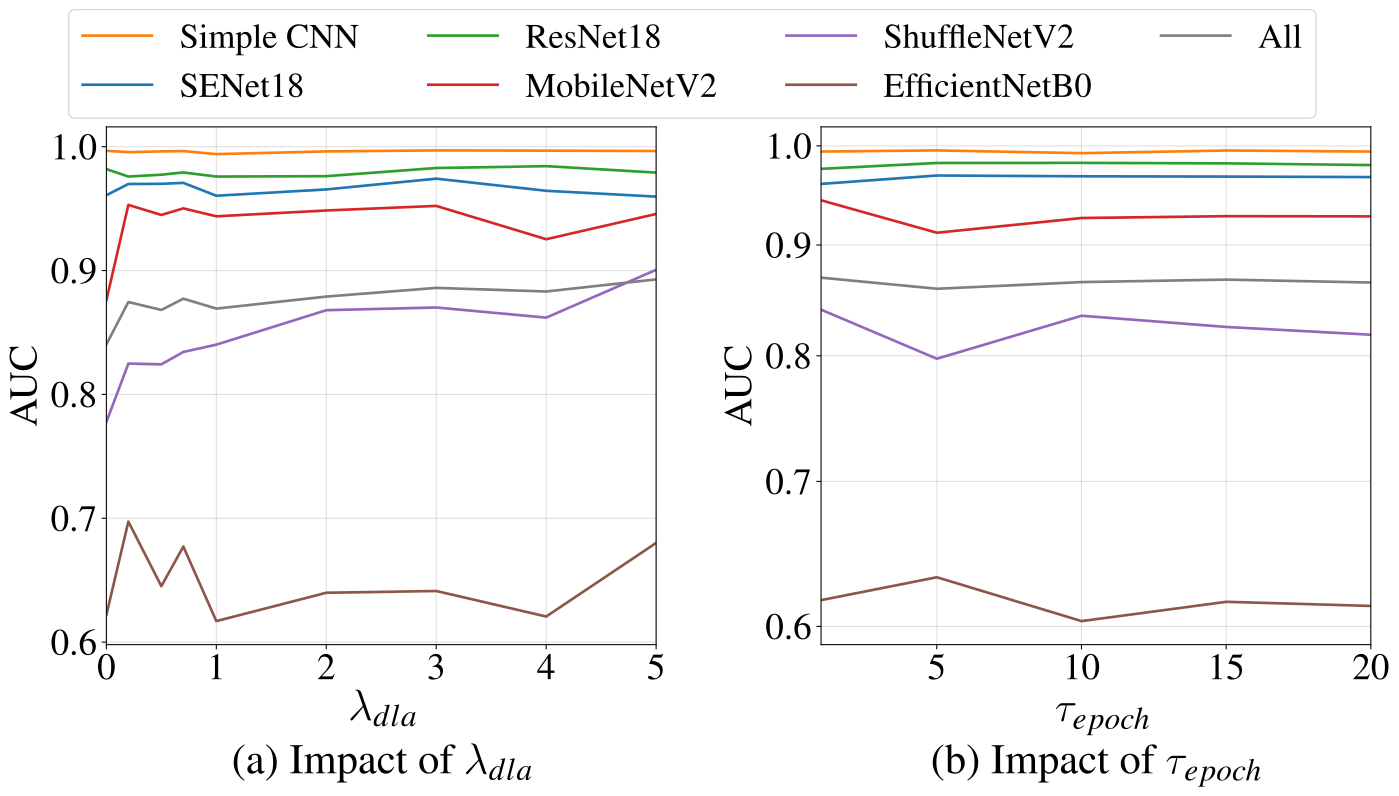}
\caption{Effect of hyper-parameters ${\lambda}_{dla}$ and $\tau_{epoch}$ on the detection AUC of \tool on GTSRB dataset.}
\label{fig:result_hyper}
\end{figure}
In this section, we analyze the sensitivity of hyper-parameters in \tool. 
In Section~\ref{sec:optimazation}, we have introduced the hyper-parameters ${\lambda}_{dla}$ and $\tau_{epoch}$, 
which play critical roles in the training process.
The former determines the weight of the distribution-level alignment loss, while the latter specifies the epochs at which the sample-level alignment interval is performed during training.
To assess their impact on detection performance, we have conducted experiments with varying settings on GTSRB dataset. Each experiment has been repeated five times, and the average results are shown in Figure \ref{fig:result_hyper}. 
\tool consistently performs well across different hyperparameter values. We observe that \tool maintained robust performance, suggesting that the distribution alignment loss does not necessitate an excessively high weight, and the sample-level alignment can be performed at long intervals.

\subsection{The Resistance to Adaptive Attacks}
We design two types of adaptive attacks to show the robustness of \tool, and they are stated as follows.
\subsubsection{Conflating Architecture \& Backdoor Features}
\label{sec:CABFadaptiveattack}
\tool{} aims to extract architecture-invariant backdoor features from models to detect backdoors effectively. This assumes the existence of such architecture-invariant backdoor features. However, the conflating architecture and backdoor features attack attempts to challenge this assumption by blurring the distinction between architecture and backdoor features. This attack involves training backdoored models such that the backdoor features strongly correlate with the model architecture, rendering our detection method ineffective.

Specifically, assuming the adversary has full knowledge of \tool's feature extractor $g_f(\cdot)$, and they train \( N \) sets of backdoored models simultaneously, each set comprising models with the same architecture \(\mathcal{S} = \{\mathcal{S}^i \mid i=1, \ldots, N\}\). During training, they use \( g_f \) to extract features from different models and aim to bring features of models with the same architecture closer in the feature space while pushing features of models with different architectures apart. 
This is achieved through the conflating loss, which is defined as:
\begin{equation} 
\begin{split}
\mathcal{L}_{con} = &\sum_{\substack{m_{k}^i, m_{l}^i \in \mathcal{S}^i \\ i \in \{1, \ldots, N\}
}} D_C(g_f(m_{k}^i), g_f(m_{l}^i)) \\ & - \sum_{\substack{m_{k}^i \in \mathcal{S}^i,m_{l}^j \in \mathcal{S}^j \\ i \neq j \in \{1, \ldots, N\}}} D_C(g_f(m_{k}^i), g_f(m_{l}^j)),
\end{split}
\end{equation}
where $D_C(\cdot,\cdot)$ represents the cosine distance.
\begin{figure}[tb]
    \centering
    \includegraphics[width=\linewidth]{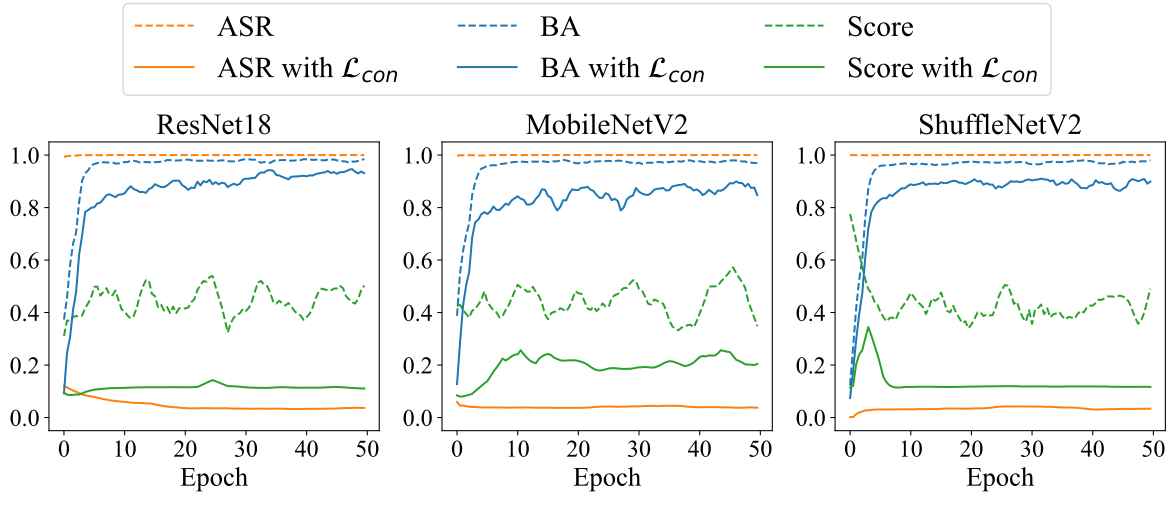}
    \caption{The influence of the $\mathcal{L}_{con}$ loss during the insertion of backdoors using the BadNets attack. Solid lines represent training with $\mathcal{L}_{con}$, while dashed lines depict training without it. BA and ASR denote the model's benign accuracy and attack success rate during training, respectively. 
    The detection score reflects \tool{}'s prediction, where higher scores indicate a higher likelihood of a model being backdoored.}
    \label{fig:ca_ada3}
\end{figure}

Experiments are conducted with five sets of models, each comprising three architectures (ResNet18, MobileNetV2, ShuffleNetV2), using the BadNets attack to insert backdoors. We compare the outcomes of backdoor insertion with and without the $\mathcal{L}_{con}$ loss, as shown in Figure~\ref{fig:ca_ada3}. 
Without the $\mathcal{L}_{con}$ loss, models achieve an ASR of 1 early in training, remaining stable as BA improves. When $\mathcal{L}_{con}$ is included during training, although BA behaves normally, the ASR consistently stays near 0, indicating a failure in backdoor insertion. The experimental results highlight the considerable challenges posed by conflating architecture and backdoor features, suggesting that most backdoor features are architecture-invariant. 

\subsubsection{Adversarial Purge Attacks on \tool}
We propose another adaptive attack, the adversarial purge attack, which seeks to minimize the disparity between \tool's outputs for backdoored and benign models by fine-tuning target backdoor models. 

We consider a powerful adversary with complete knowledge of our detection method, including all parameters of the detection model and the query set $\mathbf{q}$. Additionally, they can access all training data used to construct any backdoored model $m_t$ and can train a benign model $m_b$ without a backdoor. The adversary attempts to bypass \tool at three different stages:

\noindent
\textit{Purging Backdoored Model Output.} The adversary aims to make the output $\mathbf{p}$ of the backdoored model $m_t$ on the query set $\mathbf{q}$ as consistent as possible with the benign model $m_b$ with the same queries. To achieve this, the attacker fine-tunes $m_t$ with an additional purge loss $\mathcal{L}_{purgep}$:
\begin{equation} 
\mathcal{L}_{purgep} = D_C(m_t(\mathbf{q}), m_b(\mathbf{q})),
\end{equation}
where $D_C(\cdot,\cdot)$ represents the cosine distance. 

\noindent
\textit{Purging Features.} The adversary aims to make the features $\mathbf{e}$ extracted by \tool from the backdoored model $m_t$ as similar as possible to those extracted from the benign model $m_b$. To achieve this, the adversary fine-tunes $m_t$ with an additional purge loss $\mathcal{L}_{purgee}$:
\begin{equation} 
\mathcal{L}_{purgee} = D_C(g_f(m_t), g_f(m_b)),
\end{equation}
where $g_f(\cdot)$ denotes the feature extractor of \tool.

\noindent
\textit{Purging Detection Outcome.} The adversary attempts to make the detection model classify the backdoored model $m_t$ as a benign model. To achieve this, the adversary fine-tunes $m_t$ with a purge loss $\mathcal{L}_{purget}$:
\begin{equation} 
\mathcal{L}_{purget} = \ell_{l1}(g(m_t), g(m_b)),
\end{equation}
where $\ell_{l1}(\cdot,\cdot)$ denotes the $L1$ loss.

\begin{table*}[t]
\caption{\change{Detection AUC under Adversarial Purge Attacks (\tool{} vs. MNTD)}}
\label{tab:ad_auc}
\begin{tabular}{ccccccccc}
\toprule
\multirow{2}{*}{\textbf{Adversarial Purge Attacks}} & \multirow{2}{*}{\textbf{Approach}} & \multicolumn{3}{c}{\textbf{Given Model Architectures}} & \multicolumn{3}{c}{\textbf{Unseen Model Architectures}} & \multirow{2}{*}{All} \\ \cmidrule(lr){3-8}
 &  & Simple CNN & SENet18 & ResNet18 & MobileNetV2 & ShuffeNetV2 & EfficientNetB0 &  \\ \midrule
Purging Backdoored Model Output & \tool & 0.9885 & 0.9462 & 0.9737 & 0.8888 & 0.8115 & 0.6128 & 0.8722 \\
Purging Features & \tool & 0.9296 & 0.5824 & 0.7074 & 0.6594 & 0.7047 & 0.6086 & 0.7136 \\
Purging Detection Outcome & \tool & 0.5416 & 0.8151 & 0.8391 & 0.7158 & 0.5822 & 0.5759 & 0.6860 \\
\begin{tabular}[c]{c}Purging Backdoored Model Output\\ (Model Output as Features)\end{tabular} & MNTD & 0.6042 & 0.5445 & 0.5094 & 0.5433 & 0.5816 & 0.5304 & 0.6026 \\
Purging Detection Outcome & MNTD & 0.4870 & 0.6321 & 0.5149 & 0.5625 & 0.5613 & 0.5004 & 0.3877 \\ \bottomrule
\end{tabular}
\end{table*}

\begin{figure}[tb]
    \centering
    \includegraphics[width=\linewidth]{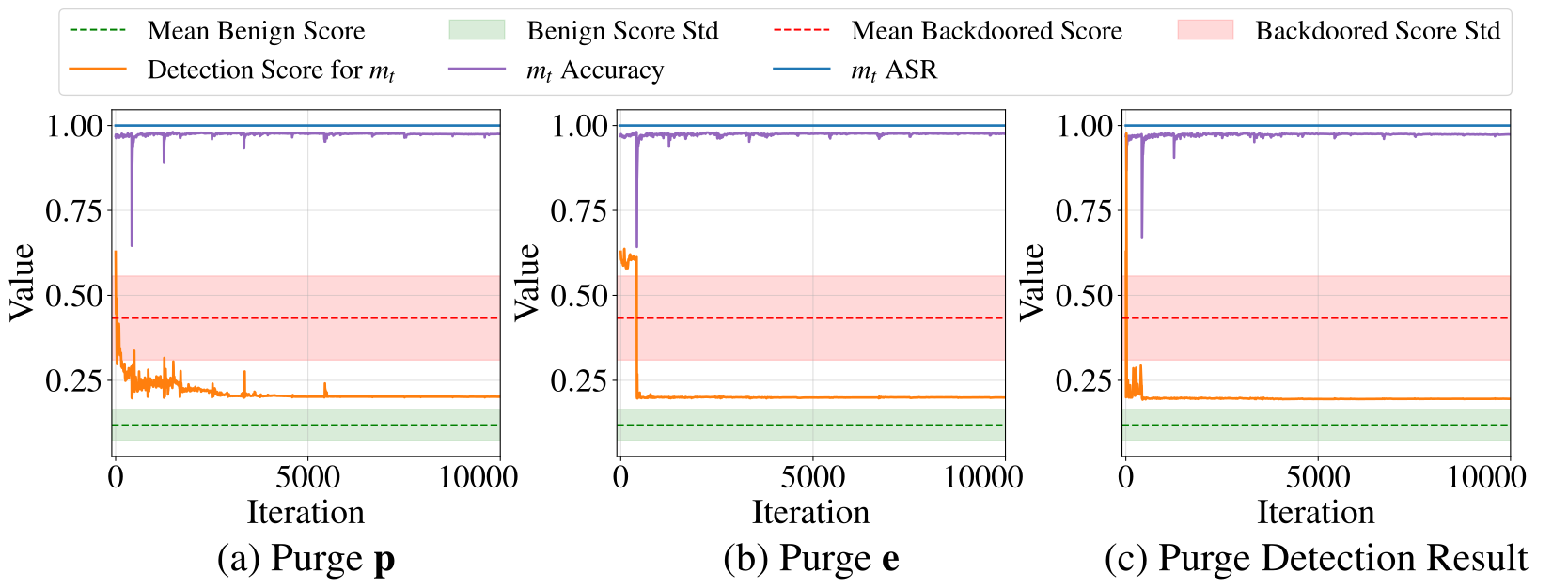}
    \caption{The impact of (a) $\mathcal{L}_{purgep}$, (b) $\mathcal{L}_{purgee}$, and (c) $\mathcal{L}_{purget}$ on the benign accuracy, ASR, and detection scores of \tool for the backdoored model $m_t$ \change{(MobileNetV2, GTSRB)}. We also include the mean and standard deviation of scores for both the benign and backdoored models.}
    \label{fig:adaptive_attacks}
\end{figure}

Figure~\ref{fig:adaptive_attacks} illustrates how each purging strategy affects the BA, ASR, and detection score of $m_t$. Although the detection scores can be partially suppressed, $m_t$ consistently struggles to fully fall within the benign distribution range, especially under $\mathcal{L}_{purget}$. This demonstrates the robustness of \tool{} against fine-tuning-based evasion.

We hypothesize that the robustness of our method stems from the nature of the query set. 
It contains various trigger information that activates backdoor features when the target model retains them during training.
Removing its backdoors is the only way to ensure the model does not possess these backdoor-related features.

\change{\textbf{Detection AUC Under Adversarial Purge Attacks.}
To quantitatively evaluate \tool's robustness, we report detection AUC under each purge variant in Table~\ref{tab:ad_auc}, and compare against MNTD. Results show that while attacks reduce AUC, \tool{} consistently outperforms MNTD across all seen and unseen architectures, highlighting its stronger resilience under adaptive attacks.}

\change{\subsection{Computational Cost and Efficiency Analysis}
In this section, we analyze the computational overhead of our method, including the training cost for proxy models and the efficiency of the detector compared to representative baselines such as MNTD and ULPs. }

\change{\subsubsection{Training Models} To enable architecture generalization, \tool{} requires training of a diverse set of benign and proxy backdoored models. For each dataset, we train models under three architectures (Simple CNN, ResNet-18, and SENet-18), with 256 benign and 256 backdoored models per architecture. Training is conducted on eight NVIDIA GeForce RTX 4090 GPUs, and the average training time per model is 15 minutes. The total cost is:
\begin{equation} 
\begin{split}
 & \frac{15\mathrm{\ min/model} \times 256\mathrm{\ models} \times 3\mathrm{\ architectures} \times 2}{60\mathrm{\ min/h} \times 8\mathrm{\ GPUs}} \\ & = 48\mathrm{\ hours}.
\end{split}
\end{equation}
We emphasize that these models are shared across all target architectures, avoiding the repeated model training required by prior methods such as MNTD and ULPs when evaluating new architecture's target models.}

\change{\subsubsection{\tool{} Training Cost}
The primary training cost of \tool{} comes from optimizing the detector using adversarial learning and the joint loss function involving the target, distribution-level alignment, and sample-level alignment losses. As shown in Table~\ref{tab:training_cost} we compare the training time of \tool{} with MNTD and ULPs under the GTSRB dataset using the same architecture setup.
Although \tool{} incurs a 6–7× longer training time than MNTD, the training is performed only once and generalizes to unseen architectures and tasks. In contrast, baselines require retraining for each target architecture, leading to higher cumulative training costs as the number of target models increases.}

\begin{table}[t]
    \centering
    \caption{\change{Comparison of the training cost on GTSRB.}}
    \label{tab:training_cost}
    \begin{tabular}{cccc}
        \toprule
        \textbf{Method} & \textbf{Epochs} & \textbf{Time/Epoch (s)} & \textbf{Total Time (h)} \\
        \midrule
        \tool{} & 300 & 42.5 & 3.54 \\
        MNTD    & 50  & 37.2 & 0.52 \\
        ULPs    & 500 & 41.8 & 5.81 \\
        \bottomrule
    \end{tabular}
\end{table}

\change{\subsubsection{Inference Efficiency} At inference time, all methods—including MNTD, ULPs, and \tool{}—follow a similar procedure: querying the target model to obtain logits and passing them through the detector. \tool{} introduces alignment layers composed of two fully connected layers, which adds negligible latency.}

\section{Limitation and Future Work}
\tool{} shows promising potential for detecting backdoor attacks across diverse architectures. However, similar to prior learning-based detection approaches such as MNTD and ULPs, it inherits certain limitations. In particular, \tool{} requires training with a fixed number of classes, limiting its flexibility when applied to models targeting tasks with different class configurations. In such cases, retraining the detector is necessary. Enabling detectors to generalize across tasks with varying class numbers remains an open challenge for future research.

Moreover, our findings reveal that simply increasing the number of architectures seen during training does not always improve generalization. Understanding training architectures interaction effects—possibly by analyzing intermediate alignment features—offers an important research direction to further improve detector generalization and interpretability.

\section{Conclusion}
Backdoor attacks pose a substantial threat to the security and reliability of deep learning models. While existing learning-based detection solutions show promising results to alleviate this issue, their generalization ability on new model architectures hinders their wide adoption. To overcome this challenge, this paper introduces \tool, which focuses on obtaining architecture-invariant model features for robust backdoor detection. Specifically, \tool incorporates an alignment layer in the feature extraction process, utilizing two alignment losses to ensure the alignment of features from models with similar backdoors but distinct architectures. Evaluation results on 16,896 models demonstrate the superior performance of \tool compared to state-of-the-art methods. 

\section*{Acknowledgments}
This work is supported by the National Natural Science Foundation of China (Grant No. 62306093) and sponsored by the CCF-Huawei Populus Grove Fund. It is also supported by the National Key R\&D Program of China (2024YFE0215300), the National Natural Science Foundation of China (Grant No. 62376074), the Shenzhen Science and Technology Program (Grants: SGDX20230116091244004, JSGGKQTD20221101115655027, ZDSYS20230626091203008), and the Natural Science Foundation of Shandong Province (ZR2023QF131).

%

\bibliographystyle{IEEEtran}

\bibliography{ref}

\newpage

\section{Biography Section}
 
\vspace{-33pt}

\begin{IEEEbiography}[{\includegraphics[width=1in,height=1.25in,clip,keepaspectratio]{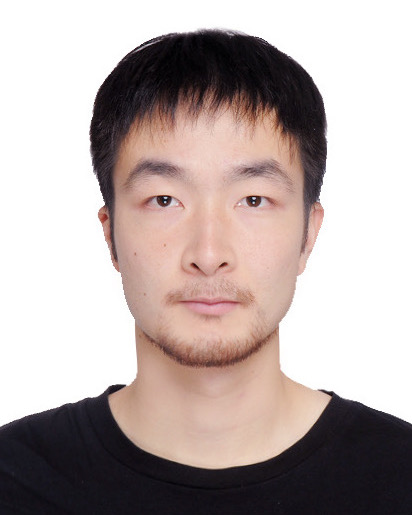}}]{Zhonghao Yang}
received the BEng degree from Northeastern University (P. R. China). He is currently pursuing the Ph.D. degree at East China Normal University. His research interests include  AI security and network and systems security.
\end{IEEEbiography}

\vspace{-33pt}
\begin{IEEEbiography}[{\includegraphics[width=1in,height=1.25in,clip,keepaspectratio]{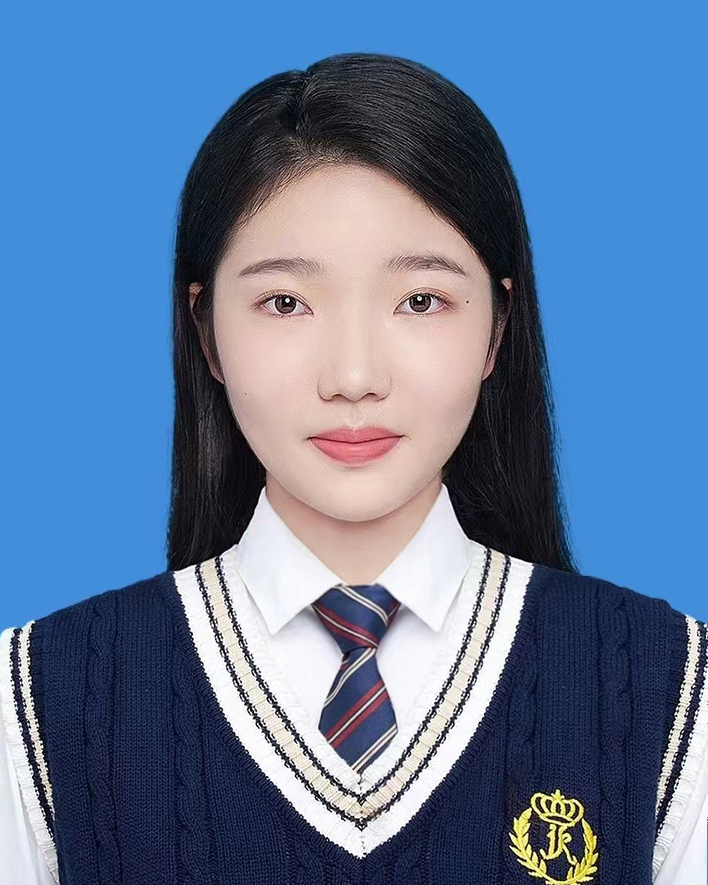}}]{Luo Cheng} is currently a postgraduate student majoring in Computer Science and Technology in Harbin Institute of Technology, Shenzhen. Her research interests are the AI safety issues. She obtained her bachelor's degree from Northeastern University in 2023. 
\end{IEEEbiography}
\vspace{-33pt}

\begin{IEEEbiography}[{\includegraphics[width=1in,height=1.25in,clip,keepaspectratio]{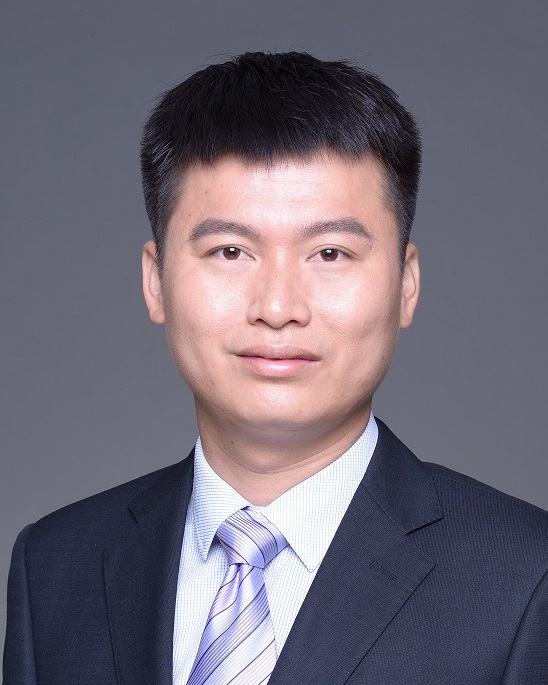}}]{Daojing He} received the B.Eng.(2007) and M. Eng. (2009) degrees from Harbin Institute of Technology (China) and the Ph.D. degree (2012) from Zhejiang University (China), all in Computer Science. He is currently a professor in the School of Computer Science and Technology, Harbin Institute of Technology, Shenzhen, China. His research interests include network and systems security. He is on the editorial board of some international journals such as IEEE Communications Magazine.
\end{IEEEbiography}

\vspace{-33pt}
\begin{IEEEbiography}[{\includegraphics[width=1in,height=1.25in,clip,keepaspectratio]{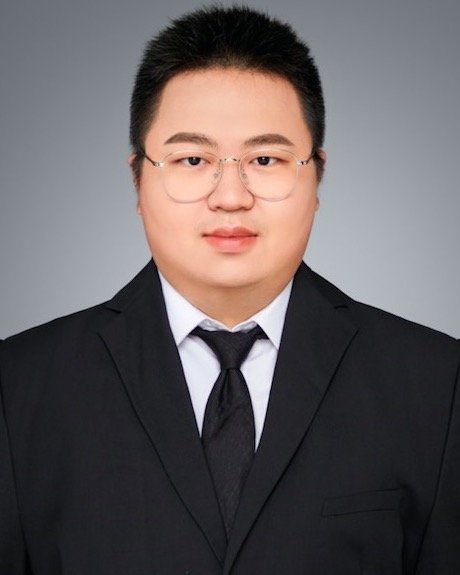}}]{Yiming Li}
 is currently a Research Fellow at Nanyang Technological University. Before that, he was a Research Professor in the State Key Laboratory of Blockchain and Data Security at Zhejiang University and also in HIC-ZJU. He received his Ph.D. degree with honors in Computer Science and Technology from Tsinghua University in 2023 and his B.S. degree with honors in Mathematics from Ningbo University in 2018. His research interests are in the domain of Trustworthy ML and Responsible AI, especially backdoor learning and AI copyright protection. His research has been published in multiple top-tier conferences and journals, such as S\&P, NDSS, ICML, CVPR, and IEEE TIFS. He served as the Area Chair of ACM MM, the Senior Program Committee Member of AAAI, and the Reviewer of IEEE TPAMI, IEEE TIFS, IEEE TDSC, etc. His research has been featured by major media outlets, such as IEEE Spectrum. He was the recipient of the Best Paper Award at PAKDD (2023), the Rising Star Award at WAIC (2023), the KAUST Rising Stars in AI (2024), and the DAAD AInet Fellowship (2024).
\end{IEEEbiography}

\vspace{-33pt}
\begin{IEEEbiography}[{\includegraphics[width=1in,height=1.25in,clip,keepaspectratio]{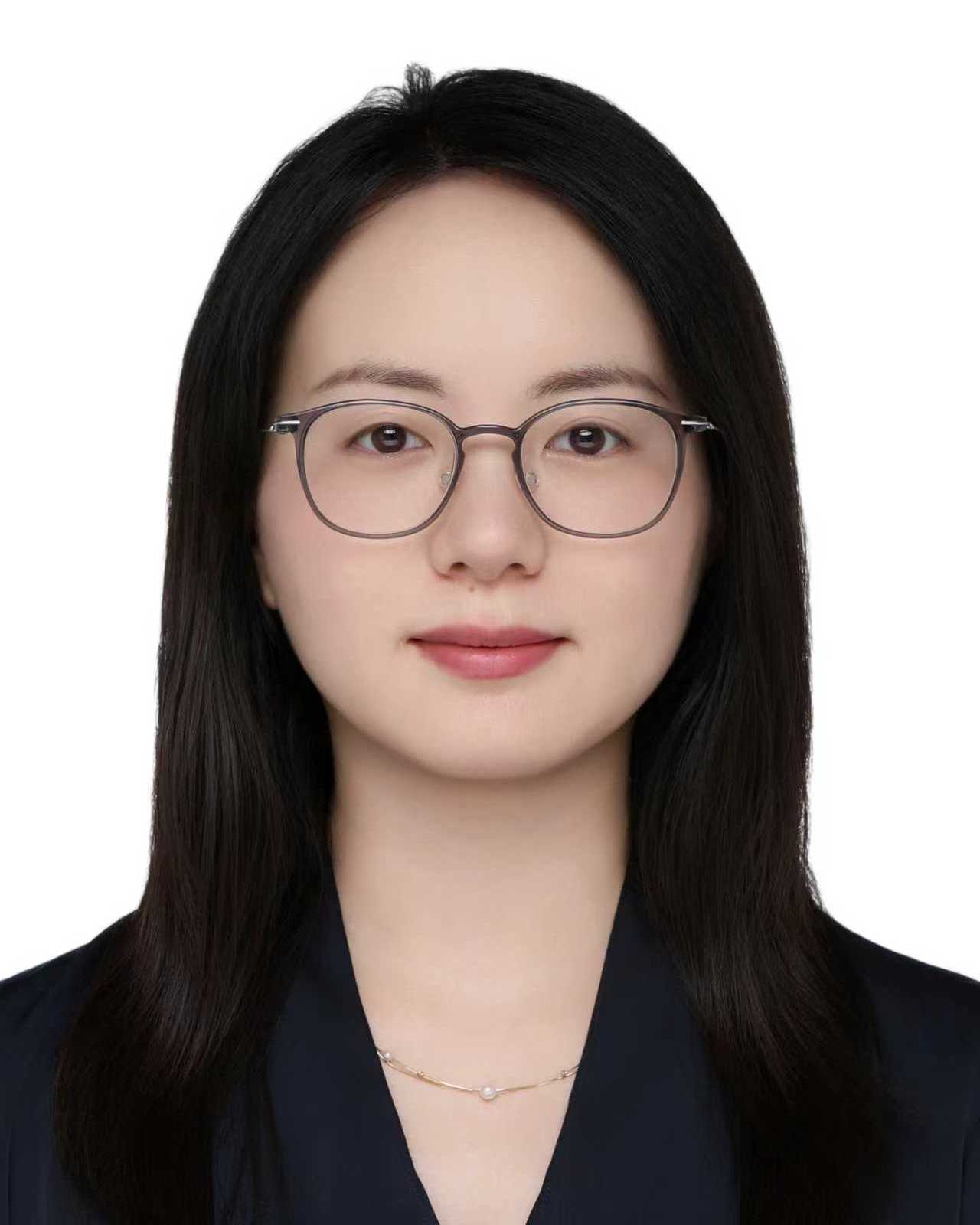}}]{Yu Li} is currently serving as a ZJU100 Professor at the School of Integrated Circuits, Zhejiang University. She received her Ph.D. degree from the Department of Computer Science and Engineering at the Chinese University of Hong Kong in 2022, obtained her master’s degree from the Katholieke Universiteit Leuven (KU Leuven) in 2017, and obtained her bachelor’s degree from the KU Leuven and the University
of Electronic Science and Technology (UESTC) in
2016. Li Yu’s main research direction is AI security
and testing. She has published many papers at leading conferences like CCS, NDSS, NeurIPS, ICML, and ISSTA.
She was nominated as a candidate for the 2022 Young Scholar Ph.D. Dissertation Award at The Chinese University of Hong Kong. In recognition of her outstanding work, she received the Best Ph.D. Dissertation Award at the 2022 Asian Test Symposium and the runner-up of the E. J. McCluskey Ph.D. Dissertation Award by the IEEE Test Technology Technical Council (TTTC).
\end{IEEEbiography}

\vfill

\end{document}